\documentclass[10pt,twocolumn,letterpaper]{article}
\pdfoutput = 1
\usepackage{iccv}
\usepackage{times}
\usepackage{epsfig}
\usepackage{graphicx}
\usepackage{amsmath}
\usepackage{amssymb}
\usepackage{textcomp}
\usepackage{subcaption}

\usepackage[pagebackref=true,breaklinks=true,letterpaper=true,colorlinks,bookmarks=false]{hyperref}

\iccvfinalcopy 
\newcommand{\samethanks}{\footnotemark[1]}
\def\iccvPaperID{****} 

\ificcvfinal\fi
\newcommand{\ignore}[1]{}

\newcommand{\SSH}{\textit{SSH}\xspace}
\newcommand{\VGG}{\textit{VGG-16}\xspace}
\newcommand{\Wider}{\textit{WIDER}\xspace}
\begin{document}

\title{SSH: Single Stage Headless Face Detector}

\author{\ Mahyar Najibi\thanks{Authors contributed equally} 
\hspace{3ex}
 Pouya Samangouei\samethanks   \hspace{3ex} Rama Chellappa  \hspace{3ex} Larry S. Davis \\
University of Maryland\\
{\tt\small { najibi@cs.umd.edu \quad \{pouya,rama,lsd\}@umiacs.umd.edu}}
}

\maketitle
\thispagestyle{empty}

\begin{abstract}
We introduce the Single Stage Headless (\SSH) face detector. Unlike two stage proposal-classification detectors, \SSH detects faces in a single stage directly from the early convolutional layers in a classification network. \SSH is headless. That is, it is able to achieve state-of-the-art results while removing the ``head" of  its underlying classification network -- \ie all fully connected layers in the \VGG which contains a large number of parameters. Additionally, instead of relying on an image pyramid to detect faces with various scales, \SSH is scale-invariant by design. We simultaneously detect faces with different scales in a single forward pass of the network, but from different layers. These properties make \SSH fast and light-weight. Surprisingly, with a headless \VGG, \SSH beats the \textit{ResNet-101}-based state-of-the-art on the \Wider dataset. Even though, unlike the current state-of-the-art, \SSH does not use an image pyramid and is $5X$ faster. Moreover, if an image pyramid is deployed, our light-weight network achieves state-of-the-art on all subsets of the \Wider dataset, improving the \textit{AP} by $2.5\%$. \SSH also reaches state-of-the-art results on the \textit{FDDB} and \textit{Pascal-Faces} datasets while using a small input size, leading to a runtime of $50$ ms/image on a GPU. The code is available at \href{https://github.com/mahyarnajibi/SSH}{\url{https://github.com/mahyarnajibi/SSH}}.

\end{abstract}

\vspace{-3px}
\section{Introduction}
\label{sec:intro}
Face detection is a crucial step in various problems involving verification, identification, expression analysis, \etc. From the Viola-Jones \cite{viola} detector to recent work by Hu \etal \cite{tiny}, the performance of face detectors has been improved dramatically. However, detecting small faces is still considered a challenging task. The recent introduction of the \textit{WIDER} face dataset \cite{wider}, containing a large number of small faces, exposed the performance gap between humans and current face detectors. The problem becomes more challenging when the speed and memory efficiency of the detectors are taken into account. The best performing face detectors are usually slow and have high memory foot-prints (\eg \cite{tiny} takes more than $1$ second to process an image, see Section \ref{sec:timing}) partly due to the huge number of parameters as well as the way robustness to scale or incorporation of context are addressed.

\begin{figure}[t]	
    \centering
        \includegraphics[width=\linewidth]{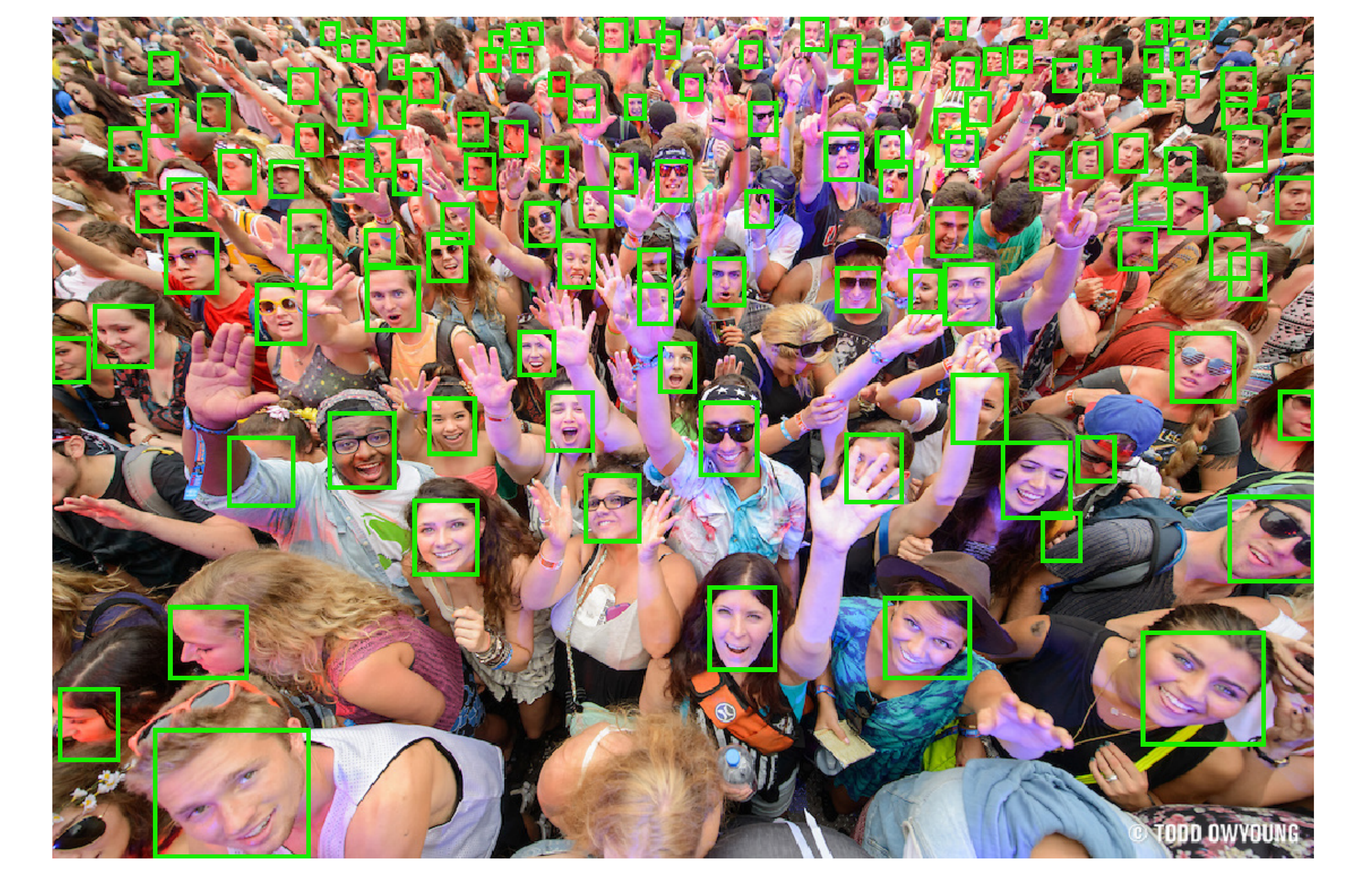}
 	\caption{\SSH  detects various face sizes in a single CNN forward pass and without employing an image pyramid in $\sim 0.1$ second for an image with size $800\times1200$ on a GPU.}
    \label{fig:confusion}
    \vspace{-10px}
\end{figure}

State-of-the-art CNN-based detectors convert image classification networks into two-stage detection systems \cite{fast,ren2015faster}. In the first stage, early convolutional feature maps are used to propose a set of candidate object boxes. In the second stage, the remaining layers of the classification networks (\eg \textit{fc6\texttildelow{}8} in \VGG \cite{simonyan2014very}), which we refer to as the network ``\textit{head}'', are deployed to extract local features for these candidates and classify them. The head in the classification networks can be computationally expensive (\eg the network head contains $\sim 120$M  parameters in \VGG and $\sim12$M parameters in \textit{ResNet-101}). Moreover, in the two stage detectors, the computation must be performed for all proposed candidate boxes. 

Very recently, Hu \etal \cite{tiny} showed state-of-the-art results on the \Wider face detection benchmark by using a similar approach to the Region Proposal Networks (\textit{RPN}) \cite{ren2015faster} to directly detect faces. Robustness to input scale is achieved by introducing an image pyramid as an integral part of the method. However, it involves processing an input pyramid with an up-sampling scale up to $5000$ pixels per side and passing each level to a very deep network which increased inference time.

In this paper, we introduce the Single Stage Headless (\SSH) face detector. \SSH performs detection in a single stage. Like \textit{RPN} \cite{ren2015faster}, the early feature maps in a classification network are used to regress a set of predefined anchors towards faces. However, unlike two-stage detectors, the final classification takes place together with regressing the anchors. \SSH is headless. It is able to achieve state-of-the-art results while removing the head of its underlying network (\ie all fully connected layers in \VGG), leading to a light-weight detector. Finally, \SSH is scale-invariant by design. Instead of relying on an external multi-scale pyramid as input, inspired by \cite{lin2016fpn}, \SSH detects faces from various depths of the underlying network. This is achieved by placing an efficient convolutional detection module on top of the layers with different strides, each of which is trained for an appropriate range of face scales. Surprisingly, \SSH based on a headless \VGG, not only outperforms the best-reported \VGG by a large margin but also beats the current \textit{ResNet-101}-based state-of-the-art method on the \Wider face detection dataset. Unlike the current state-of-the-art, \SSH does not deploy an input pyramid and is $5$ times faster. If an input pyramid is used with \SSH as well, our light-weight \VGG-based detector outperforms the best reported \textit{ResNet-101} \cite{tiny} on all three subsets of the \Wider dataset and improves the mean average precision by $4\%$ and $2.5\%$ on the validation and the test set respectively. \SSH also achieves state-of-the-art results on the \textit{FDDB} and \textit{Pascal-Faces} datasets with a relatively small input size, leading to a runtime of $50$ ms/image.

The rest of the paper is organized as follows. Section \ref{sec:related} provides an overview of the related works. Section \ref{sec:ssh} introduces the proposed method. Section \ref{sec:exp} presents the experiments and Section \ref{sec:conclusion} concludes the paper.
\section{Related Works}
\label{sec:related}
\subsection{Face Detection}
Prior to the re-emergence of convolutional neural networks (\textit{CNN}), different machine learning algorithms were developed to improve face detection performance \cite{viola,zhu2012face,li2013probabilistic,li2014efficient,mathias2014face,chen2014joint,yang2014aggregate}. However, following the success of these networks in classification tasks \cite{krizhevsky2012imagenet}, they were applied to detection as well \cite{girshick2014rich}. Face detectors based on \textit{CNN}s significantly closed the performance gap between human and artificial detectors \cite{li2015convolutional,yang2015facial,yang2015convolutional,cms,tiny}. However, the introduction of the challenging \Wider dataset \cite{wider}, containing a large number of small faces, re-highlighted this gap. To improve performance, \textit{CMS-RCNN} \cite{cms} changed the \textit{Faster R-CNN} object detector \cite{ren2015faster} to incorporate context information. Very recently, Hu \etal  proposed a face detection method based on proposal networks which achieves state-of-the-art results on this dataset \cite{tiny}. However, in addition to skip connections, an input pyramid is processed by re-scaling the image to different sizes, leading to slow detection speeds. In contrast, \SSH is able to process multiple face scales simultaneously in a single forward pass of the network, which reduces inference time noticeably.

\subsection{Single Stage Detectors and Proposal Networks}
The idea of detecting and localizing objects in a single stage has been previously studied for general object detection. \textit{SSD} \cite{liu2016ssd} and \textit{YOLO} \cite{redmon2016you} perform detection and classification simultaneously by classifying a fixed grid of boxes and regressing them towards objects. \textit{G-CNN} \cite{najibi2016g} models detection as a piece-wise regression problem and iteratively pushes an initial multi-scale grid of boxes towards objects while classifying them. However, current state-of-the-art methods on the challenging \textit{MS-COCO} object detection benchmark are based on two-stage detectors\cite{lin2014microsoft}. \SSH is a single stage detector; it detects faces directly from the early convolutional layers without requiring a proposal stage.

Although \SSH is a detector, it is more similar to the object proposal algorithms which are used as the first stage in detection pipelines. These algorithms generally regress a fixed set of \textit{anchors} towards objects and assign an objectness score to each of them. \textit{MultiBox} \cite{szegedy2014scalable} deploys clustering to define anchors. \textit{RPN} \cite{ren2015faster}, on the other hand, defines anchors as a dense grid of boxes with various scales and aspect ratios, centered at every location in the input feature map. \SSH uses similar strategies, but to localize and at the same time detect, faces.

\subsection{Scale Invariance and Context Modeling}
Being scale invariant is important for detecting faces in unconstrained settings. For generic object detection,  \cite{bell2016inside,zagoruyko2016multipath} deploy feature maps of earlier convolutional layers to detect small objects. Recently, \cite{lin2016fpn} used skip connections in the same way as \cite{long2015fully} and employed multiple shared \textit{RPN} and classifier heads from different convolutional layers. For face detection, \textit{CMS-RCNN} \cite{cms} used the same idea as \cite{bell2016inside,zagoruyko2016multipath} and added skip connections to the \textit{Faster RCNN} \cite{ren2015faster}. \cite{tiny} creates a pyramid of images and processes each separately to detect faces of different sizes. In contrast, \SSH is capable of detecting faces at different scales in a single forward pass of the network without creating an image pyramid. We employ skip connections in a similar fashion as \cite{long2015fully,lin2016fpn}, and train three detection modules jointly from the convolutional layers with different strides to detect small, medium, and large faces.

In two stage object detectors, context is usually modeled by enlarging the window around proposals \cite{zagoruyko2016multipath}. \cite{bell2016inside} models context by deploying a recurrent neural network. For face detection, \textit{CMS-RCNN} \cite{cms} utilizes a larger window with the cost of duplicating the classification head. This increases the memory requirement as well as detection time. \SSH uses simple convolutional layers to achieve the same larger window effect, leading to more efficient context modeling.
\section{Proposed Method}
\label{sec:ssh}
\SSH is designed to decrease inference time, have a low memory foot-print, and be scale-invariant. \SSH is a single-stage detector; \ie instead of dividing the detection task into bounding box proposal and classification, it performs classification together with localization from the global information extracted from the convolutional layers. We empirically show that in this way, \SSH can remove the ``head'' of its underlying network while achieving state-of-the-art face detection accuracy. Moreover, \SSH is scale-invariant by design and can incorporate context efficiently.

\subsection{General Architecture}
\begin{figure*}
    \centering
    \includegraphics[width=0.8\linewidth]{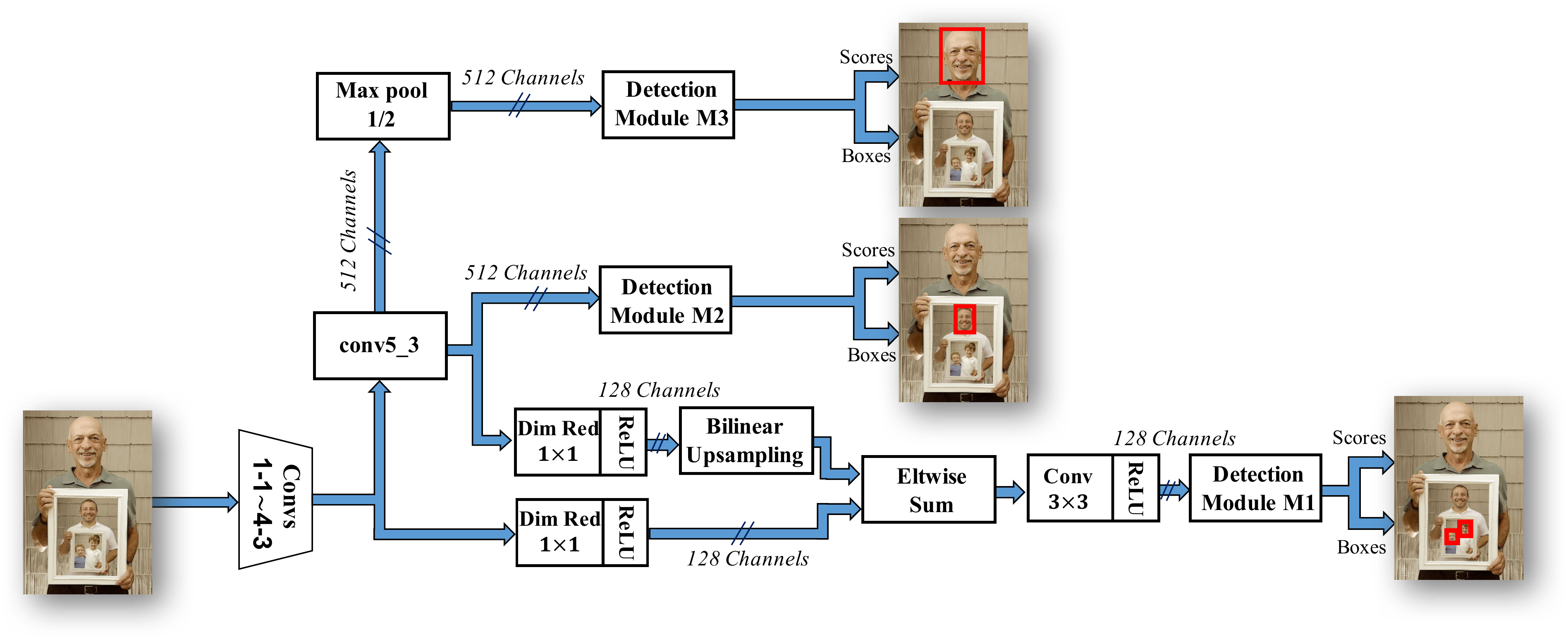}
 	\caption{The network architecture of \SSH.}
    \label{fig:general_arch}
\end{figure*}

Figure \ref{fig:general_arch} shows the general architecture of \SSH. It is a fully convolutional network which localizes and classifies faces early on by adding a \textit{detection module} on top of feature maps with strides of $8$, $16$, and $32$, depicted as $\mathcal{M}_1$, $\mathcal{M}_2$, and $\mathcal{M}_3$ respectively. The \textit{detection module} consists of a convolutional binary classifier and a regressor for detecting faces and localizing them respectively. 

To solve the localization sub-problem, as in \cite{szegedy2014scalable,ren2015faster,najibi2016g}, \SSH regresses a set of predefined bounding boxes called \textit{anchor}s, to the ground-truth faces. We employ a similar strategy to the \textit{RPN} \cite{ren2015faster} to form the anchor set. We define the anchors in a dense overlapping sliding window fashion. At each sliding window location, $K$ anchors are defined which have the same center as that window and different scales. However, unlike \textit{RPN}, we only consider anchors with aspect ratio of one to reduce the number of anchor boxes. We noticed in our experiments that having various aspect ratios does not have a noticeable impact on face detection precision. More formally, if the feature map connected to the detection module $\mathcal{M}_i$ has a size of $W_i\times H_i$, there would be $W_i \times H_i \times K_i$ anchors with aspect ratio one and scales $\{S_i^1, S_i^2, \hdots S_i^{K_i}\}$. 

\begin{figure}
    \centering
    \includegraphics[width=0.6\linewidth]{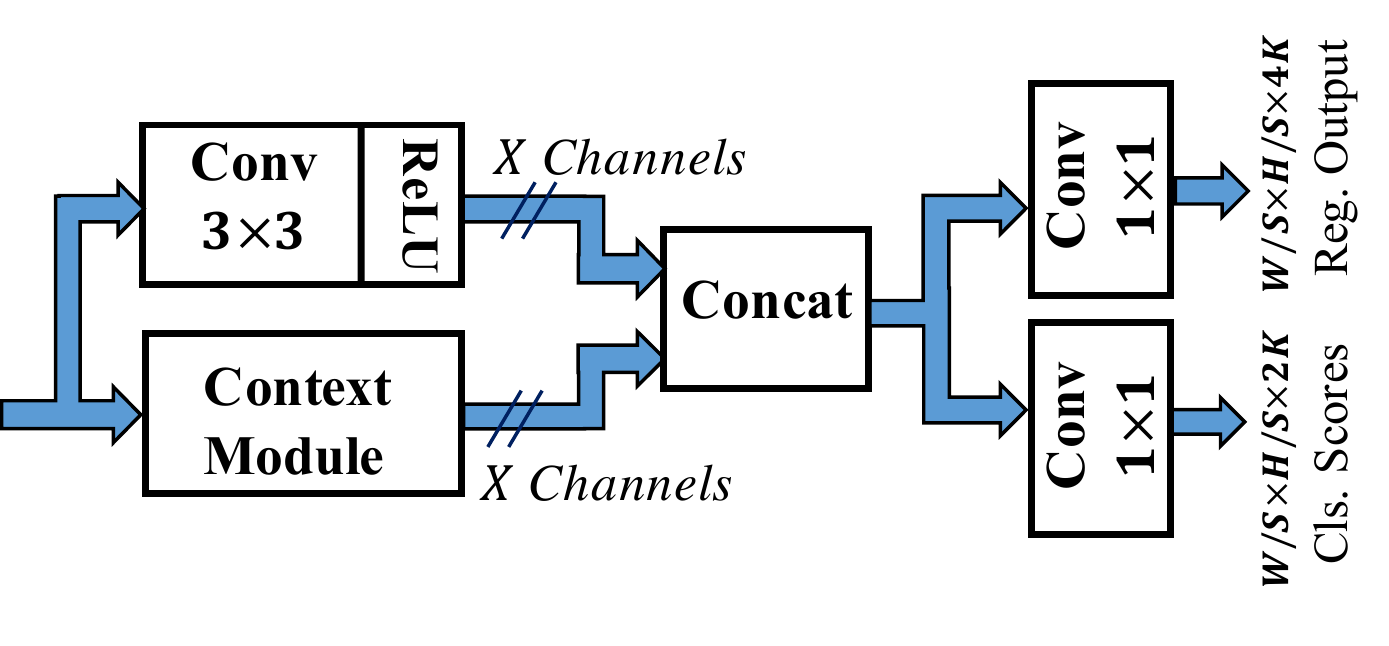}
 	\caption{\SSH detection module.}
    \label{fig:det_module}
\end{figure}

For the detection module, a set of convolutional layers are deployed to extract features for face detection and localization as depicted in Figure \ref{fig:det_module}. This includes a simple context module to increase the effective receptive field as discussed in section \ref{sec:context}. The number of output channels of the context module, (\ie ``$X$'' in Figures \ref{fig:det_module} and \ref{fig:context}) is set to $128$ for detection module $\mathcal{M}_1$ and $256$ for modules $\mathcal{M}_2$ and $\mathcal{M}_3$. Finally, two convolutional layers perform bounding box regression and classification. At each convolution location in $\mathcal{M}_i$, the classifier decides whether the windows at the filter's center and corresponding to each of the scales $\{S_i^k\}_{k=1}^K$ contains a face. A $1 \times 1$ convolutional layer with $2 \times K$ output channels is used as the classifier. For the regressor branch, another $1 \times 1$ convolutional layer with $4\times K$ output channels is deployed. At each location during the convolution, the regressor predicts the required change in scale and translation to match each of the \textit{positive} anchors to faces.

\subsection{Scale-Invariance Design}
\label{sec:scale_invar}
In unconstrained settings, faces in images have varying scales. Although forming a multi-scale input pyramid and performing several forward passes during inference, as in \cite{tiny}, makes it possible to detect faces with different scales, it is slow. In contrast, \SSH detects large and small faces  simultaneously in a single forward pass of the network. Inspired by \cite{lin2016fpn}, we detect faces from three different convolutional layers of our network using detection modules $\mathcal{M}_1,\mathcal{M}_2$, and $\mathcal{M}_3$. These modules have strides of $8$, $16$, and $32$ and are designed to detect small, medium, and large faces respectively. 

More precisely, the detection module $\mathcal{M}_2$ performs detection from the \textit{conv5-3} layer in \VGG. Although it is possible to place the detection module $\mathcal{M}_1$ directly on top of \textit{conv4-3}, we use the feature map fusion which was previously deployed for semantic segmentation \cite{long2015fully}, and generic object detection \cite{lin2016fpn}. However, to decrease the memory consumption of the model, the number of channels in the feature map is reduced from 512 to 128 using $1\times1$ convolutions. The \textit{conv5-3} feature maps are up-sampled and summed up with the \textit{conv4-3} features, followed by a $3 \times 3$ convolutional layer. We used bilinear up-sampling in the fusion process. For detecting larger faces, a max-pooling layer with stride of $2$ is added on top of the \textit{conv5-3} layer to increase its stride to $32$. The detection module $\mathcal{M}_3$ is placed on top of this newly added layer.

During the training phase, each detection module $\mathcal{M}_i$ is trained to detect faces from a target scale range as discussed in \ref{sec:training}. During inference, the predicted boxes from the different scales are joined together followed by Non-Maximum Suppression (\textit{NMS}) to form the final detections. 

\subsection{Context Module}
\label{sec:context}
\begin{figure}
    \centering
    \includegraphics[width=0.7\linewidth]{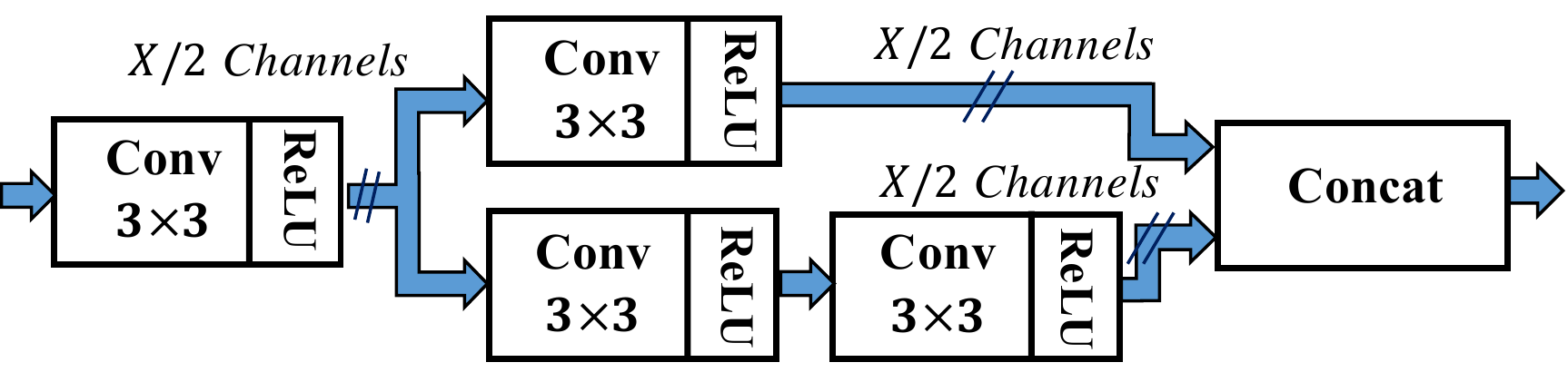}
 	\caption{\SSH context module.}
    \label{fig:context}
\end{figure}

In two-stage detectors, it is common to incorporate context by enlarging the window around the candidate proposals. \SSH mimics this strategy by means of simple convolutional layers. Figure \ref{fig:context} shows the context layers which are integrated into the detection modules. Since anchors are classified and regressed in a convolutional manner, applying a larger filter resembles increasing the window size around proposals in a two-stage detector. To this end, we use $5 \times 5$ and $7 \times 7$ filters in our context module. Modeling the context in this way increases the receptive field proportional to the stride of the corresponding layer and as a result the target scale of each detection module. To reduce the number of parameters, we use a similar approach as \cite{szegedy2015going} and deploy sequential $3 \times 3$ filters instead of larger convolutional filters. The number of output channels of the detection module (\ie ``$X$'' in Figure \ref{fig:context}) is set to $128$ for $\mathcal{M}_1$ and $256$ for modules $\mathcal{M}_2$ and $\mathcal{M}_3$. It should be noted that our detection module together with its context filters uses fewer of parameters compared to the module deployed for proposal generation in \cite{ren2015faster}. Although, more efficient, we empirically found that the context module improves the mean average precision on the \Wider validation dataset by more than half a percent.

\subsection{Training}
\label{sec:training}
We use stochastic gradient descent with momentum and weight decay for training the network. As discussed in section \ref{sec:scale_invar}, we place three detection modules on layers with different strides to detect faces with different scales. Consequently, our network has three multi-task losses for the classification and regression branches in each of these modules as discussed in Section \ref{sec:loss}. To specialize each of the three detection modules for a specific range of scales, we only back-propagate the loss for the anchors which are assigned to faces in the corresponding range. This is implemented by distributing the anchors based on their size to these three modules (\ie smaller anchors are assigned to $\mathcal{M}_1$ compared to $\mathcal{M}_2$, and $\mathcal{M}_3$). An anchor is assigned to a ground-truth face if and only if it has a higher IoU than $0.5$. This is in contrast to the methods based on \textit{Faster R-CNN} which assign to each ground-truth at least one anchor with the highest IoU. Thus, we do not back-propagate the loss through the network for ground-truth faces inconsistent with the anchor sizes of a module.

\subsubsection{Loss function}
\label{sec:loss}
\SSH has a multi-task loss. This loss can be formulated as follows:
\begin{align}
\sum_{k} \frac{1}{N^c_k} \sum_{i\in \mathcal{A}_k} &\ell_c(p_i,g_i) + \nonumber \\ 
&\lambda \sum_k \frac{1}{N^r_k} \sum_{i \in \mathcal{A}_k} \mathcal{I}(g_i=1) \ell_r(b_i,t_i) 
\end{align}
where $\ell_c$ is the face classification loss. We use standard multinomial logistic loss as $\ell_c$. The index $k$ goes over the \SSH detection modules $\mathcal{M}=\{\mathcal{M}_k\}_1^K$ and $\mathcal{A}_k$ represents the set of anchors defined in $\mathcal{M}_k$. The predicted category for the $i$'th anchor in $\mathcal{M}_k$ and its assigned ground-truth label are denoted as $p_i$ and $g_i$ respectively. As discussed in Section \ref{sec:scale_invar}, an anchor is assigned to a ground-truth bounding box if and only if it has an IoU greater than a threshold (\ie 0.5). As in \cite{ren2015faster}, negative labels are assigned to anchors with IoU less than a predefined threshold (\ie 0.3) with any ground-truth bounding box. $N^c_k$ is the number of anchors in module $\mathcal{M}_k$ which participate in the classification loss computation.

$\ell_r$ represents the bounding box regression loss. Following \cite{girshick2014rich,girshick2015fast,ren2015faster}, we parameterize the regression space with a log-space shift in the box dimensions and a scale-invariant translation and use smooth $\ell_1$ loss as $\ell_r$. In this parametrized space, $p_i$ represents the predicted four dimensional translation and scale shift and $t_i$ is its assigned ground-truth regression target for the $i$'th anchor in module $\mathcal{M}_k$. $\mathcal{I}(.)$ is the indicator function that limits the regression loss only to the positively assigned anchors, and $N_k^r=\sum_{i\in \mathcal{A}_k} I(g_i=1)$.

\subsection{Online hard negative and positive mining}
\label{sec:OHEM}
We use online negative and positive mining (\textit{OHEM}) for training \SSH as described in \cite{shrivastava2016training}. However, \textit{OHEM} is applied to each of the detection modules ($\mathcal{M}_k$) separately. That is, for each module $\mathcal{M}_k$, we select the negative anchors with the highest scores and the positive anchors with the lowest scores with respect to the weights of the network at that iteration to form our mini-batch. Also, since the number of negative anchors is more than the positives, following \cite{fast}, $25\%$ of the mini-batch is reserved for the positive anchors. As empirically shown in Section \ref{sec:ohem}, \textit{OHEM} has an important role in the success of \SSH which removes the fully connected layers out of the \VGG network. 
\section{Experiments}
\label{sec:exp}
\subsection{Experimental Setup}
\label{sec:setup}
All models are trained on $4$ GPUs in parallel using stochastic gradient descent. We use a mini-batch of $4$ images. Our networks are fine-tuned for $21K$ iterations starting from a pre-trained ImageNet classification network. Following \cite{fast}, we fix the initial convolutions up to \textit{conv3-1}. The learning rate is initially set to $0.004$ and drops by a factor of 10 after $18K$ iterations. We set momentum to $0.9$, and weight decay to $5e^{-4}$. Anchors with IoU$>0.5$ are assigned to positive class and anchors which have an IoU$<0.3$ with all ground-truth faces are assigned to the background class. For anchor generation, we use scales $\{1, 2\}$ in $\mathcal{M}_1$, $\{ 4, 8\}$ in $\mathcal{M}_2$, and $\{16,32\}$ in $\mathcal{M}_3$ with a base anchor size of $16$ pixels. All anchors have aspect ratio of one. During training, $256$ detections per module is selected for each image. During inference, each module outputs $1000$ best scoring anchors as detections and \textit{NMS} with a threshold of $0.3$ is performed on the outputs of all modules together. 

\subsection{Datasets}
\textbf{WIDER dataset\cite{wider}:} This dataset contains $32,203$ images with $393,703$ annotated faces, $158,989$ of which are in the train set, $39,496$ in the validation set and the rest are in the test set. The validation and test set are divided into ``easy'', ``medium'', and ``hard'' subsets cumulatively (\ie the ``hard'' set contains all images). This is one of the most challenging public face datasets mainly due to the wide variety of face scales and occlusion. We train all models on the train set of the \Wider dataset and evaluate on the validation and test sets. Ablation studies are performed on the the validation set (\ie ``hard'' subset).

\textbf{FDDB\cite{fddb}:} \textit{FDDB} contains 2845 images and 5171 annotated faces. We use this dataset only for testing.

\textbf{Pascal Faces\cite{pascalfaces}:}
\textit{Pascal Faces} is a subset of the \textit{Pascal VOC} dataset \cite{pascal} and contains 851 images annotated for face detection. We use this dataset only to evaluate our method.

\subsection{WIDER Dataset Result}
\label{sec:wider_res}
\begin{figure*}
\begin{subfigure}{.31\textwidth}
  \centering
  \includegraphics[width=.85\linewidth]{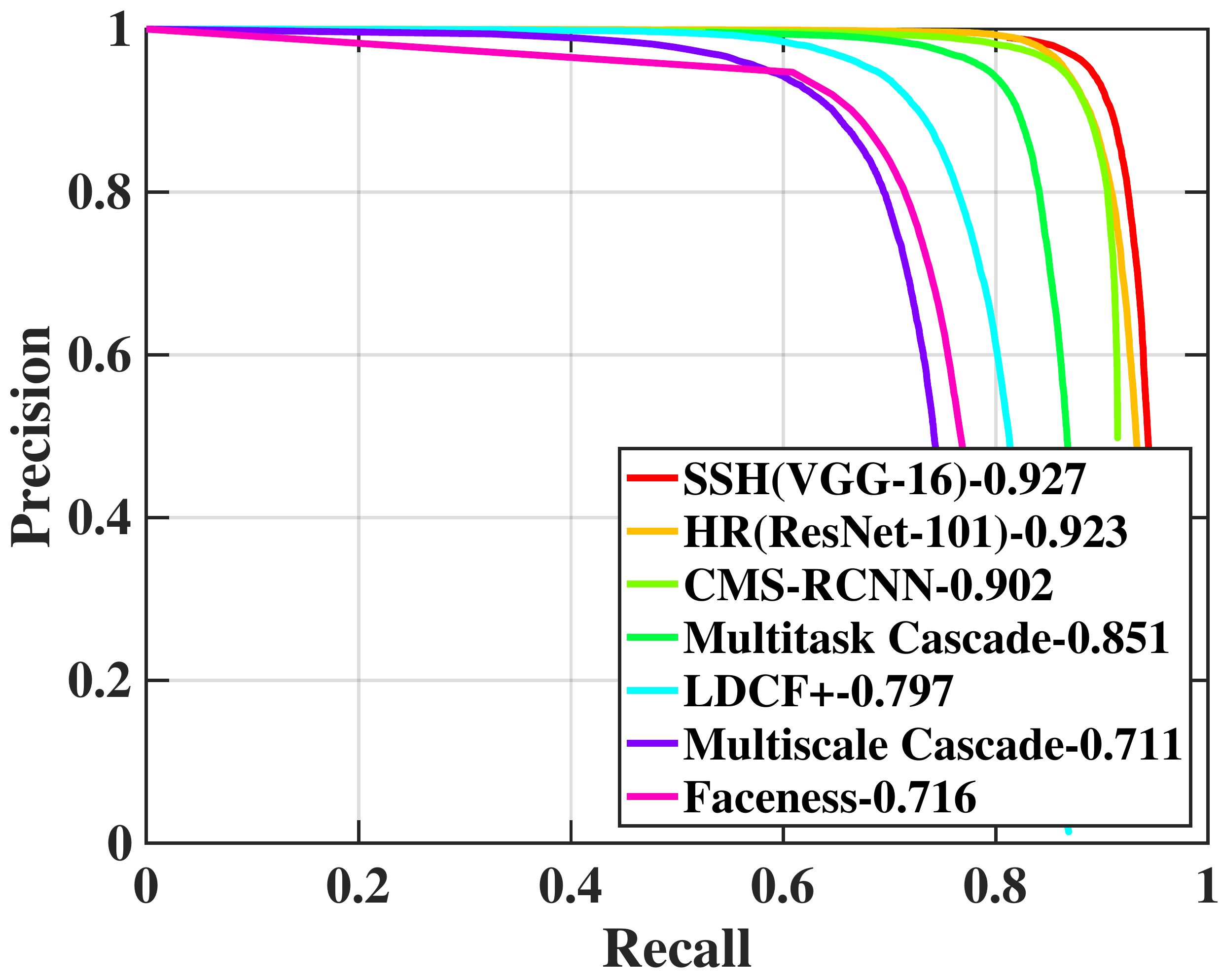}
  \caption{Easy}
  \label{fig:wider_easy}
\end{subfigure}
\begin{subfigure}{.31\textwidth}
  \centering
  \includegraphics[width=.85\linewidth]{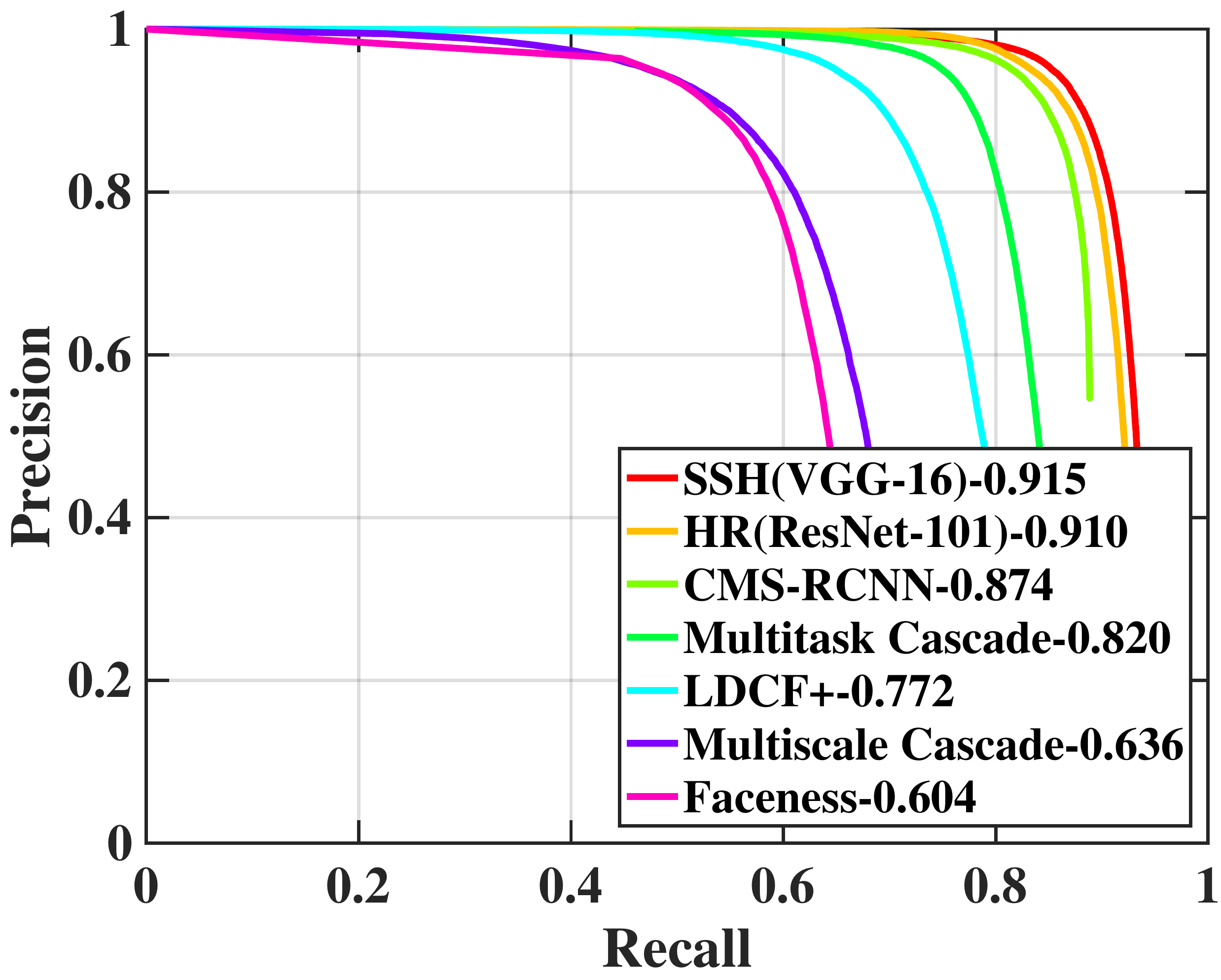}
  \caption{Medium}
  \label{fig:wider_med}
\end{subfigure}
\begin{subfigure}{.31\textwidth}
  \centering
  \includegraphics[width=.85\linewidth]{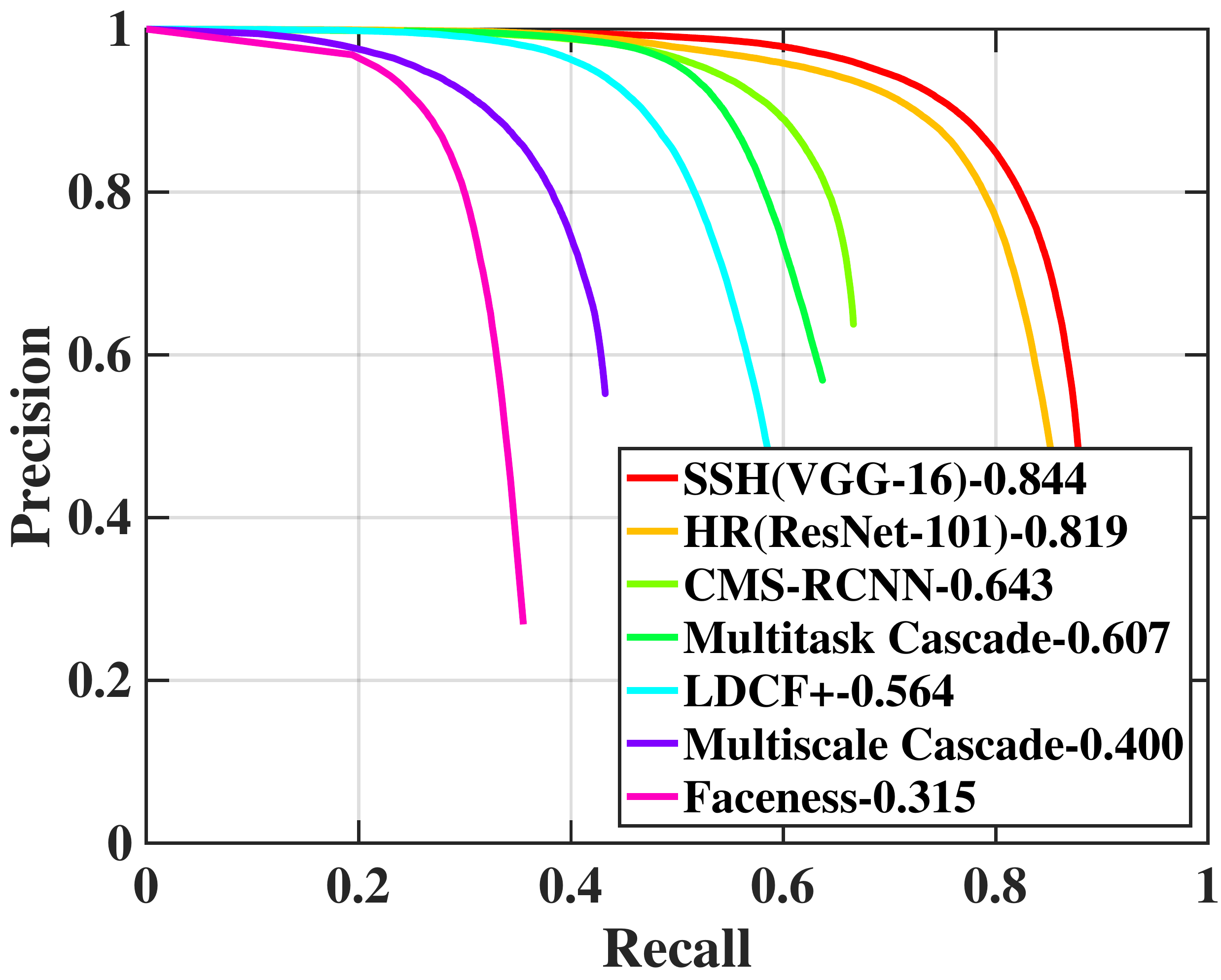}
  \caption{Hard}
  \label{fig:wider_hard}
\end{subfigure}
\caption{Comparison among the methods on the test set of \Wider face detection benchmark.}
\label{fig:wider_test}

\end{figure*}
We compare \SSH with \textit{HR} \cite{tiny}, \textit{CMS-RCNN} \cite{cms}, \textit{Multitask Cascade CNN} \cite{zhang2016joint}, \textit{LDCF} \cite{ohn2017boost}, Faceness \cite{faceness}, and \textit{Multiscale Cascade CNN} \cite{wider}. When reporting \SSH without an image pyramid, we rescale the shortest side of the image up to $1200$ pixels while keeping the largest side below $1600$ pixels without changing the aspect ratio. \textit{SSH+Pyramid} is our method when we apply \SSH to a pyramid of input images. Like \textit{HR}, a four level image pyramid is deployed. To form the pyramid, the image is first scaled to have a shortest side of up to $800$ pixels and the longest side less than $1200$ pixels. Then, we scale the image to have min sizes of $500,800,1200$, and $1600$ pixels in the pyramid. All modules detect faces on all pyramid levels, except $\mathcal{M}_3$ which is not applied to the largest level.

Table \ref{tab:HR_res} compares \SSH with best performing methods on the \Wider validation set. \SSH without using an image pyramid and based on the \VGG network outperforms the \VGG version of \textit{HR} by $5.7\%, 6.3\%$, and $6.5\%$ in ``easy'', ``medium'', and ``hard'' subsets respectively. Surprisingly, \SSH also outperforms \textit{HR} based on \textit{ResNet-101} on the whole dataset (\ie ``hard'' subset) by $0.8$. In contrast \textit{HR} deploys an image pyramid. Using an image pyramid, \SSH based on a light \VGG model, outperforms the \textit{ResNet-101} version of \textit{HR} by a large margin, increasing the state-of-the-art on this dataset by $\sim 4\%$.

The precision-recall curves on the \textit{test} set is presented in Figure \ref{fig:wider_test}. We submitted the detections of \SSH with an image pyramid only once for evaluation. As can be seen, \SSH based on a headless \VGG, outperforms the prior methods on all subsets, increasing the state-of-the-art by $2.5\%$.

\begin{table}
  \centering
  \footnotesize
  \caption{\small{Comparison of \SSH with top performing methods on the validation set of the \Wider dataset.}}
    \begin{tabular}{|c|ccc|}
    \hline
    {Method} & easy & medium & hard\\
    \hline
    CMS-RCNN \cite{cms} & 89.9& 87.4 & 62.9 \\
    \hline
    HR(VGG-16)+Pyramid \cite{tiny} & 86.2& 84.4 & 74.9 \\
	\hline
    HR(ResNet-101)+Pyramid \cite{tiny}  & 92.5 & 91.0 & 80.6 \\
    \hline
    SSH(VGG-16) & 91.9 & 90.7 & 81.4 \\
    SSH(VGG-16)+Pyramid & \textbf{93.1} & \textbf{92.1} & \textbf{84.5} \\
    \hline
    \end{tabular}%
  \label{tab:HR_res}%
\end{table}

\subsection{FDDB and Pascal Faces Results}
\label{sec:fddb_res}

\begin{figure*}

\begin{subfigure}{.33\textwidth}
  \centering
  \includegraphics[width=.9\linewidth]{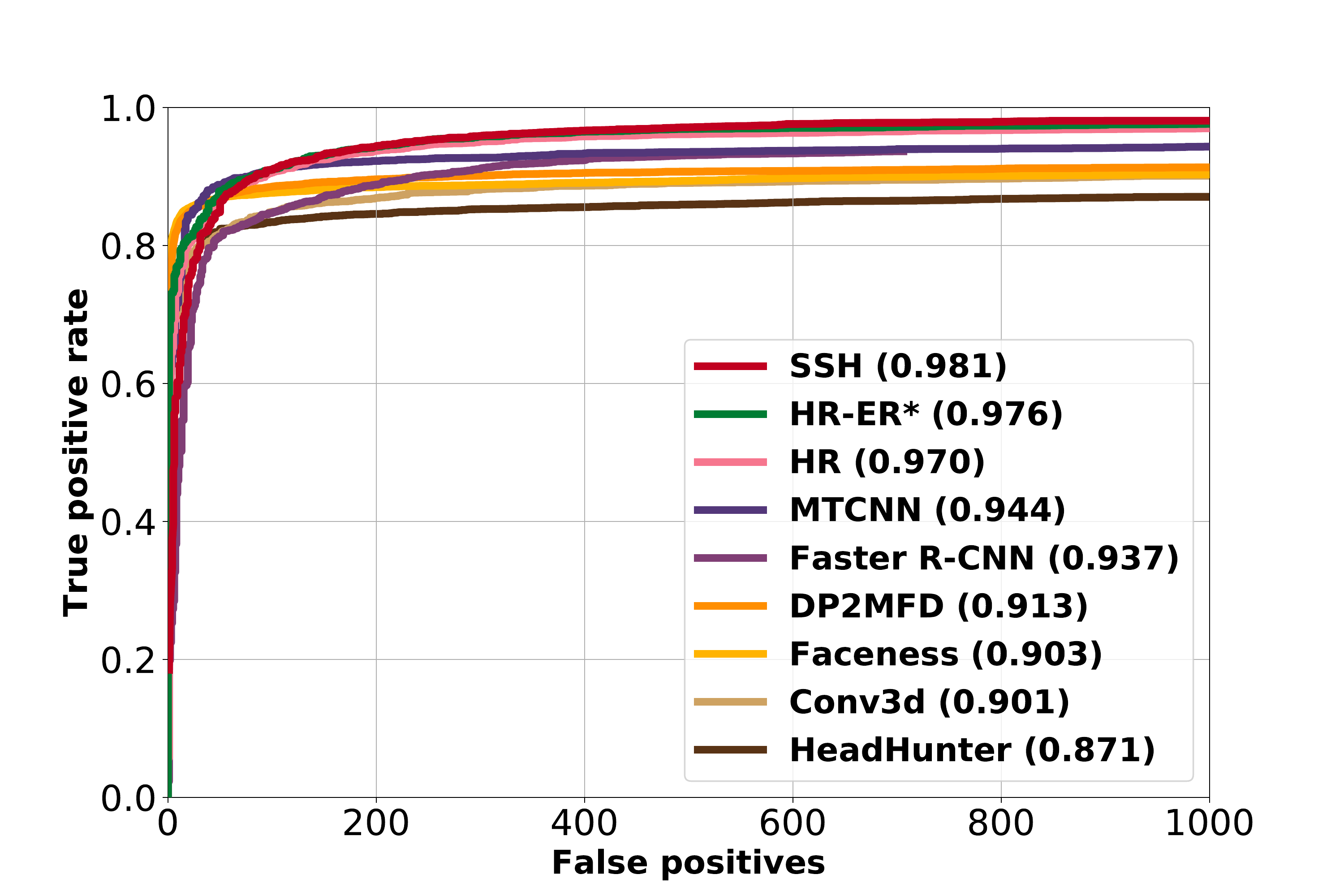}
  \caption{FDDB discrete score.}
  \label{fig:fddb_disc}
\end{subfigure}
\begin{subfigure}{.33\textwidth}
  \centering
  \includegraphics[width=.9\linewidth]{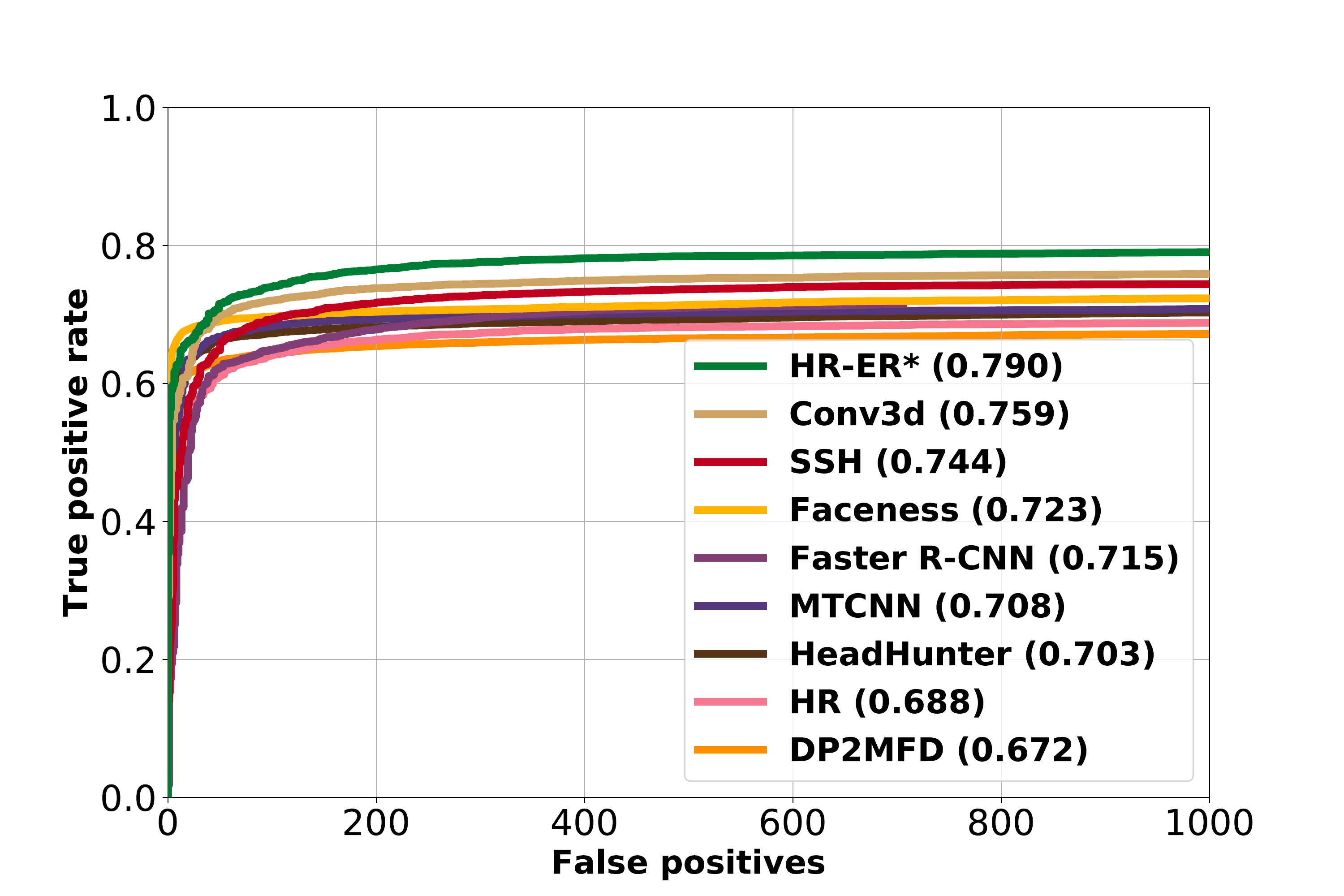}
  \caption{FDDB continuous score.}
  \label{fig:fddb_cont}
\end{subfigure}
\begin{subfigure}{.33\textwidth}
  \centering
  \includegraphics[width=.9\linewidth]{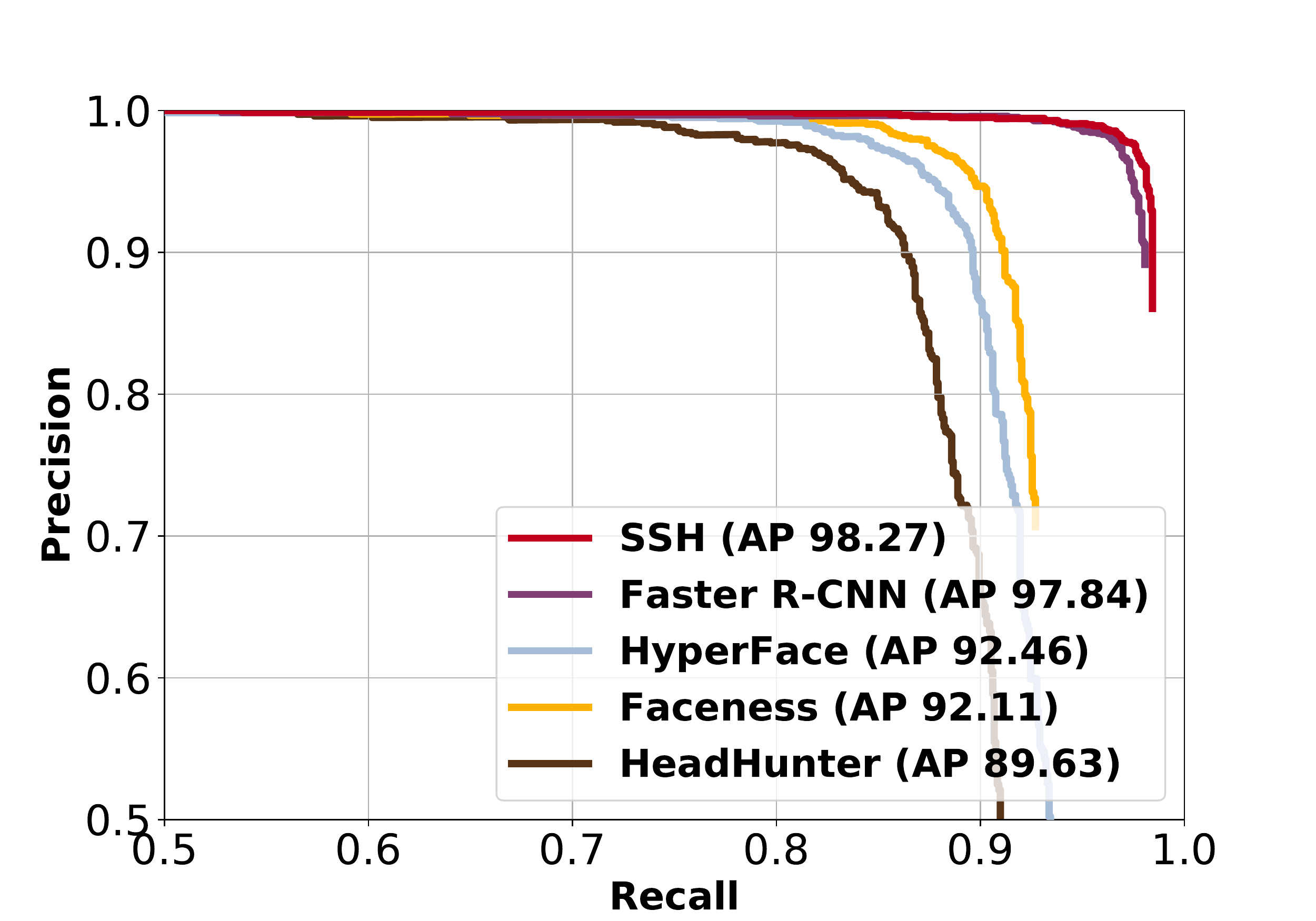}
  \caption{Pascal-Faces.}
  \label{fig:pascal}
\end{subfigure}

\caption{Comparison among the methods on FDDB and Pascal-Faces datasets. (*Note that unlike \SSH, \textit{HR-ER} is also trained on the FDDB dataset in a $10$-\textit{Fold Cross Validation} fashion.)}
\label{fig:fddb_results}
\end{figure*}

In these datasets, we resize the shortest side of the input to 400 pixels while keeping the larger side less than 800 pixels, leading to an inference time of less than $50$ ms/image. We compare \SSH with \textit{HR}\cite{tiny}, \textit{HR-ER}\cite{tiny},  Conv3D\cite{conv3d}, Faceness\cite{faceness}, Faster R-CNN(\VGG)\cite{ren2015faster}, MTCNN\cite{zhang2016joint}, DP2MFD\cite{dp2mfd}, and Headhunter\cite{mathias2014face}. Figures \ref{fig:fddb_disc} and \ref{fig:fddb_cont} show the ROC curves with respect to the discrete and continuous measures on the \textit{FDDB} dataset respectively. 

It should be noted that \textit{HR-ER} also uses \textit{FDDB} as a training data in a $10$-fold cross validation fashion. Moreover, \textit{HR-ER} and \textit{Conv3D} both generate ellipses to decrease the localization error. In contrast, \SSH does not use \textit{FDDB} for training, and is evaluated on this dataset out-of-the-box by generating bounding boxes. However, as can be seen, \SSH outperforms all other methods with respect to the discrete score. Compare to \textit{HR}, \SSH improved the results by $5.6\%$ and $1.1\%$ with respect to the continuous and discrete scores. 

We also compare \textit{SSH} with \textit{Faster R-CNN}(\VGG)\cite{ren2015faster}, \textit{HyperFace}\cite{hyperface}, \textit{Headhunter}\cite{mathias2014face}, and \textit{Faceness}\cite{faceness} on the \textit{Pascal-Faces} dataset. As shown in Figure \ref{fig:pascal}, \SSH achieves state-of-the-art results on this dataset. 

\subsection{Timing}
\label{sec:timing}
\SSH performs face detection in a single stage while removing all fully-connected layers from the \VGG network. This makes \SSH an efficient detection algorithm. Table \ref{tab:timing} shows the inference time with respect to different input sizes. We report average time on the \Wider validation set. Timing are performed on a \textit{NVIDIA Quadro P6000} GPU. In column with max size $m\times M$, the shortest side of the images are resized to ``$m$'' pixels while keeping the longest side less than ``$M$'' pixels. As shown in section \ref{sec:wider_res}, and \ref{sec:fddb_res}, \SSH outperforms \textit{HR} on all datasets without an image pyramid. On \Wider we resize the image to the last column and as a result detection takes $182$ ms/image. In contrast, \textit{HR} has a runtime of $1010$ ms/image, more than $5X$ slower. As mentioned in Section \ref{sec:fddb_res}, a maximum input size of $400\times800$ is enough for \SSH to achieve state-of-the-art performance on \textit{FDDB} and \textit{Pascal-Faces}, with a detection time of $48$ ms/image. If an image pyramid is used, the runtime would be dominated by the largest scale.   
\begin{table}
 \centering
 \caption{\small{\SSH inference time with respect to different input sizes.}}
 \label{tab:timing}
 \scalebox{0.82}{
 \begin{tabular}{|c|c|c|c|c|}
 \hline
\small{Max Size} & \small{$400\times 800$}& \small{$600\times1000$} & \small{$800\times 1200$}  & \small{$1200\times1600$}  \\
 \hline
\small{Time} & $48$ ms & $74$ ms & $107$ ms  & $182$ ms  \\
 \hline
\end{tabular}
}
\end{table}

\begin{figure*}
\begin{subfigure}{.24\textwidth}
  \centering
  \includegraphics[width=.85\linewidth]{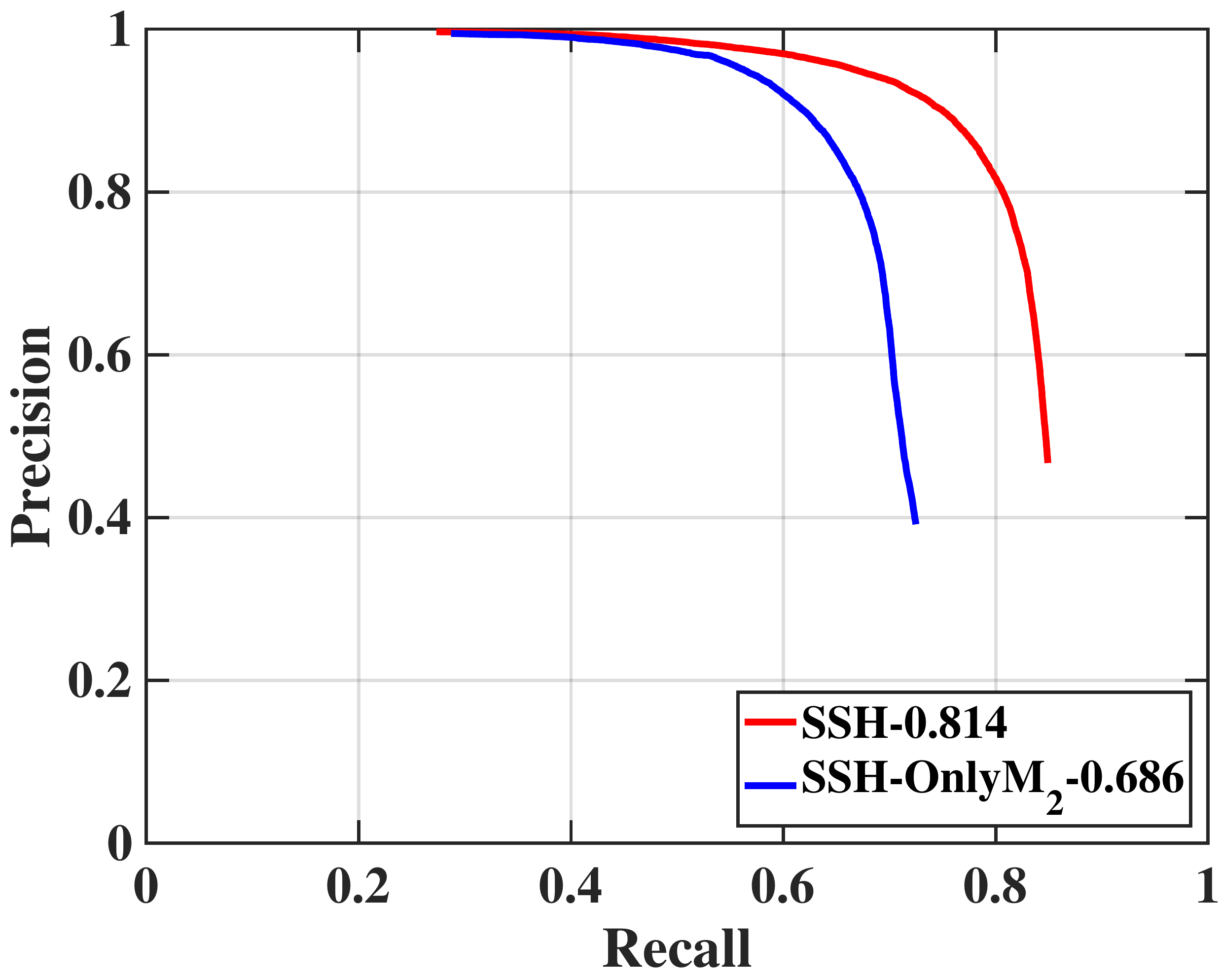}
  \caption{\small{Effect of multi-scale design.}}
  \label{fig:onlym2}
\end{subfigure}
\begin{subfigure}{.24\textwidth}
  \centering
  \includegraphics[width=.85\linewidth]{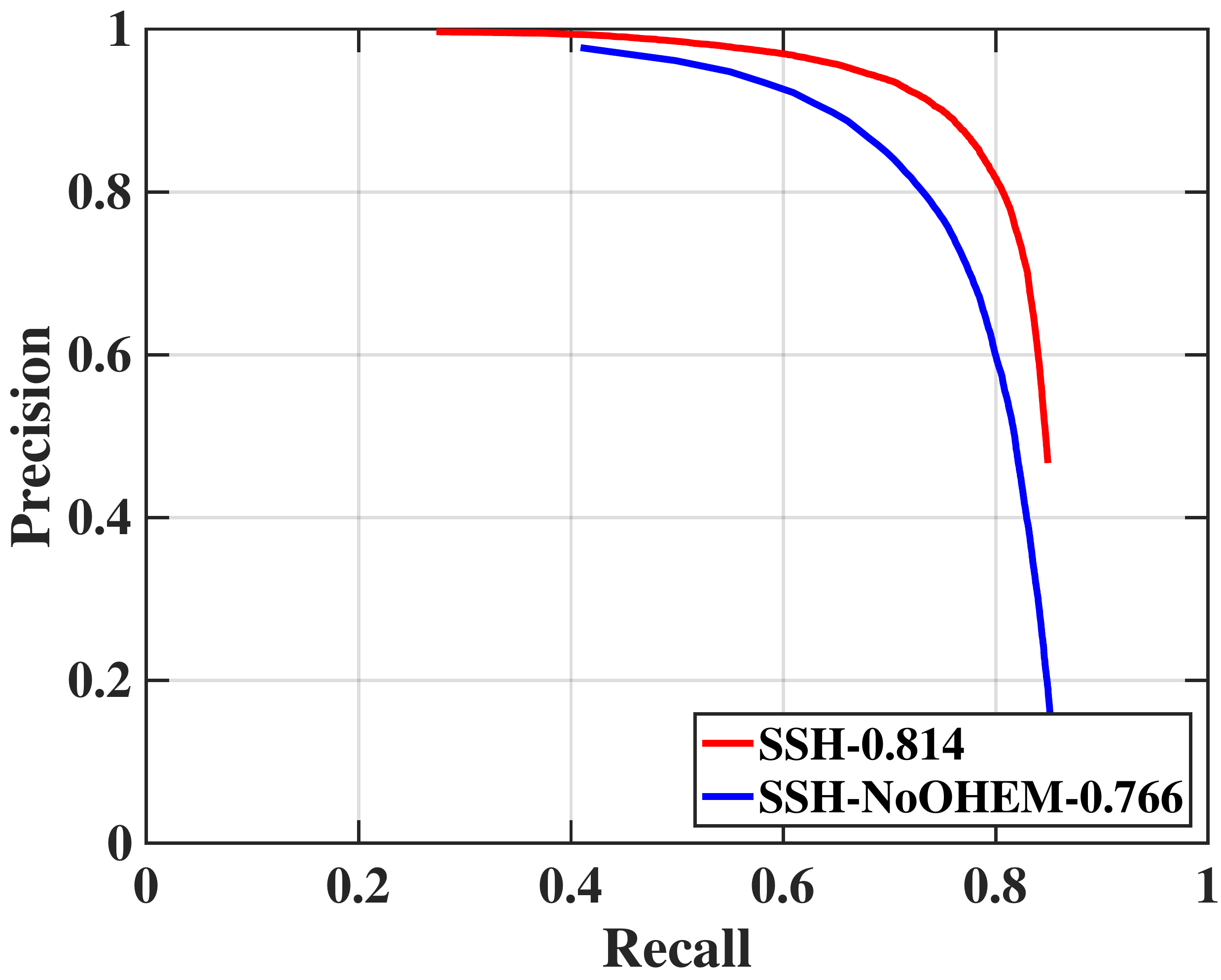}
  \caption{Effect of \textit{OHEM}.}
  \label{fig:ohem}
  
\end{subfigure}
\begin{subfigure}{.24\textwidth}
  \centering
  \includegraphics[width=.85\linewidth]{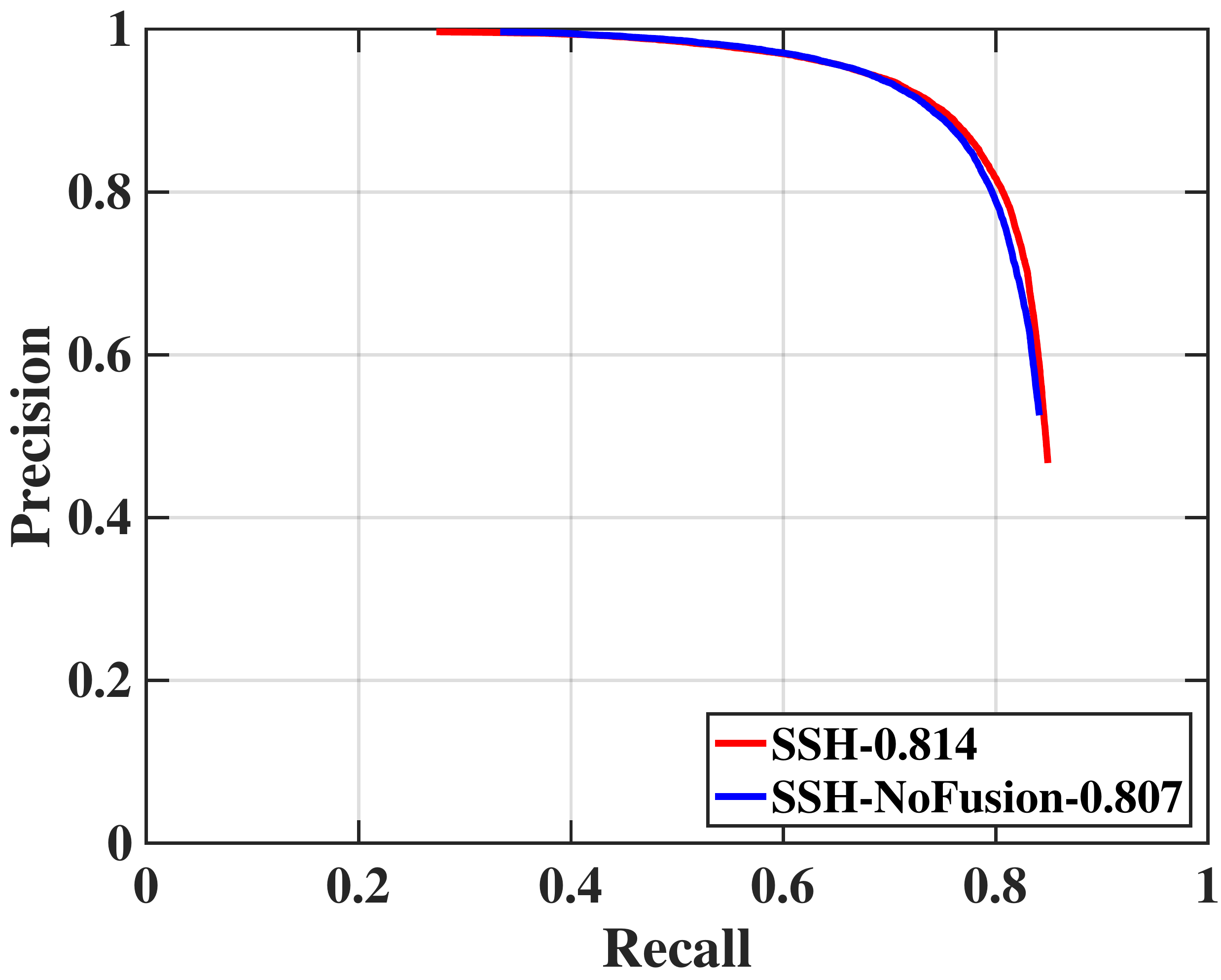}
  \caption{Effect of feature fusion.}
  \label{fig:fusion}
\end{subfigure}
\begin{subfigure}{.24\textwidth}
  \centering
  \includegraphics[width=.85\linewidth]{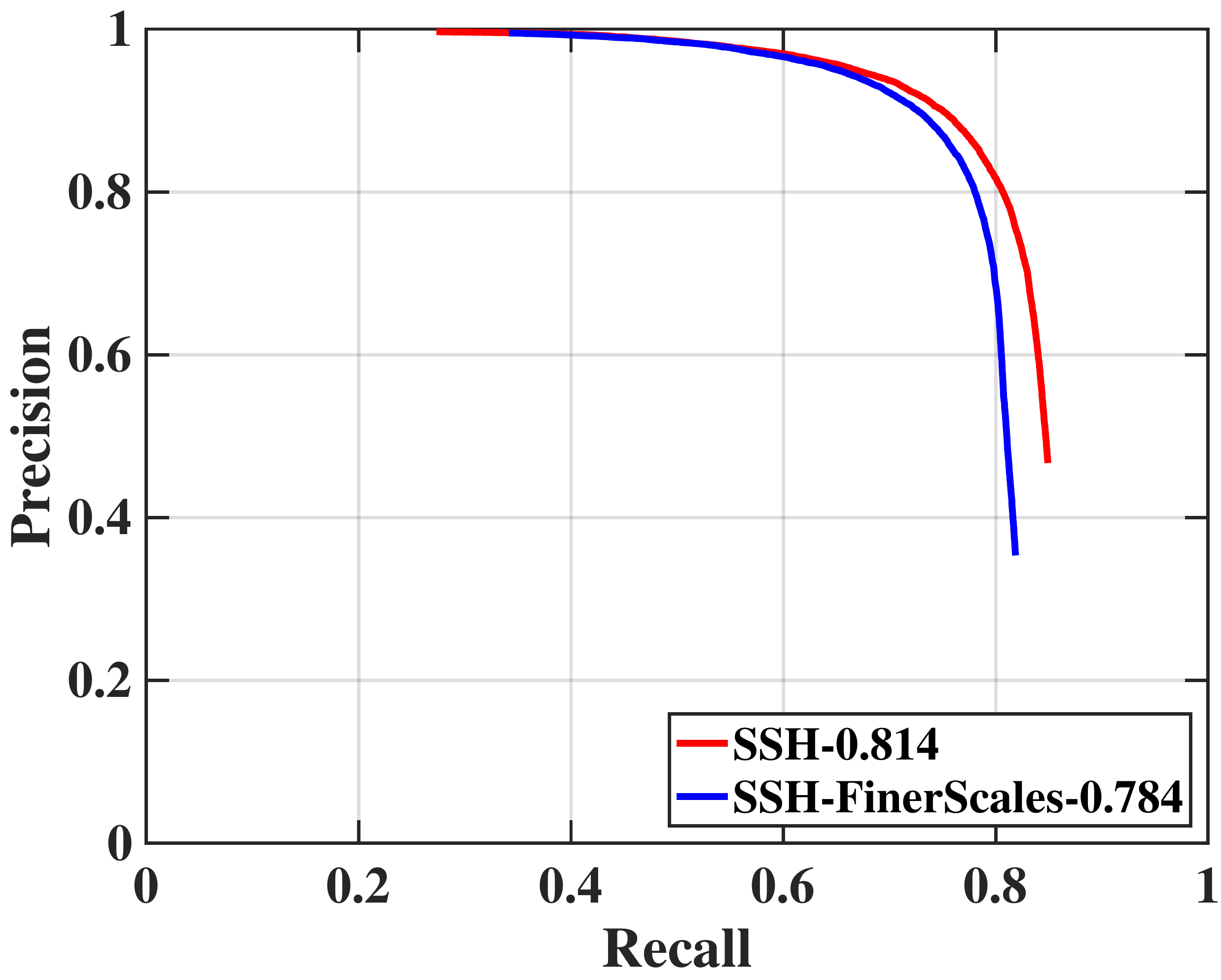}
  \caption{Effect of increasing \#anchors.}
  \label{fig:finerscales}
\end{subfigure}
\caption{Ablation studies. All experiments are reported on the Wider Validation set.} 
\label{fig:ablation}
\end{figure*}
\subsection{Ablation study: Scale-invariant design}
As discussed in Section \ref{sec:scale_invar}, \SSH uses each of its detections modules, $\{\mathcal{M}_i\}_{i=1}^3$, to detect faces in a certain range of scales from layers with different strides. To better understand the impact of these design choices, we compare the results of \SSH with and without multiple detection modules. That is, we remove $\{\mathcal{M}_1, \mathcal{M}_3\}$ and only detect faces with $\mathcal{M}_2$ from \textit{conv5-3} in \VGG. However, for fair comparison, all anchor scales in $\{\mathcal{M}_1,\mathcal{M}_3\}$ are moved to $\mathcal{M}_2$ (\ie we use $\cup_{i=1}^3 \mathbf{S}_i$ in $\mathcal{M}_2$). Other parameters remain the same. We refer to this simpler method as \textit{"SSH-Only$M_2$"}. 
As shown in Figure \ref{fig:onlym2}, by removing the multiple detection modules from \SSH, the \textit{AP} significantly drops by $\sim12.8\%$ on the \textit{hard} subset which contains smaller faces. Although \SSH does not deploy the expensive head of its underlying network, results suggest that having independent simple detection modules from different layers of the network is an effective strategy for scale-invariance.

\subsection{Ablation study: The effect of input size}
 The input size can affect face detection precision, especially for small faces. Table \ref{tab:input_size} shows the \textit{AP} of \SSH on the \Wider validation set when it is trained and evaluated with different input sizes. Even at a maximum input size of $800 \times 1200$, \SSH outperforms \textit{HR-VGG16}, which up-scales images up to $5000$ pixels, by $3.5\%$, showing the effectiveness of our scale-invariant design for detecting small faces.

\begin{table}
 \centering
 \caption{\small{The effect of input size on average precision.}}
 \label{tab:input_size}
 \scalebox{0.82}{
 \begin{tabular}{|c|c|c|c|c|}
 \hline
\small{Max Size} & \small{$600\times1000$}& \small{$800\times1200$} & \small{$1200\times1600$}  & \small{$1400\times1800$}  \\
 \hline
\small{AP} & 68.6 & 78.4  & 81.4  & 81.0  \\
 \hline
\end{tabular}
}
\end{table}

\subsection{Ablation study: The effect of OHEM}
\label{sec:ohem}
As discussed in Section \ref{sec:OHEM}, we apply hard negative and positive mining (\textit{OHEM}) to select anchors for each of our detection modules. To show its role, we train \SSH, with and without \textit{OHEM}. All other factors are the same. Figure \ref{fig:ohem} shows the results. Clearly, \textit{OHEM} is important for the success of our light-weight detection method which does not use the pre-trained head of the \VGG network.

\subsection{Ablation study: The effect of feature fusion}
In \SSH, to form the input features for detection module $\mathcal{M}_1$, the outputs of \textit{conv4-3} and \textit{conv5-3} are fused together. Figure \ref{fig:fusion}, shows the effectiveness of this design choice. Although it does not have a noticeable  computational overhead, as illustrated, it improves the \textit{AP} on the \Wider validation set.

\subsection{Ablation study: Selection of anchor scales}
As mentioned in Section \ref{sec:setup}, \SSH uses $\mathbf{S}_1 = \{1, 2\}$, $\mathbf{S}_2=\{ 4, 8\}$, $\mathbf{S}_3 = \{ 16,32\}$ as anchor scale sets. Figure \ref{fig:finerscales} compares \SSH with its slight variant which uses $\mathbf{S}_1 = \{0.25,0.5,1,2,3\}$, $\mathbf{S}_2=\{ 4,6,8,10,12\}$, $\mathbf{S}_3 = \{ 16,20,24,28,32\}$. Although using a finer scale set leads to a slower inference, it also reduces the \textit{AP} due to the increase in the number of \textit{False Positives}.

\subsection{Qualitative Results}
\begin{figure*}
\centering
\begin{tabular}{|c|c|c|c|}
\hline
\begin{minipage}{0.21\linewidth}\centering\includegraphics[trim={0cm 0cm 0cm 0cm},clip,width=1\linewidth,height=0.72\linewidth]{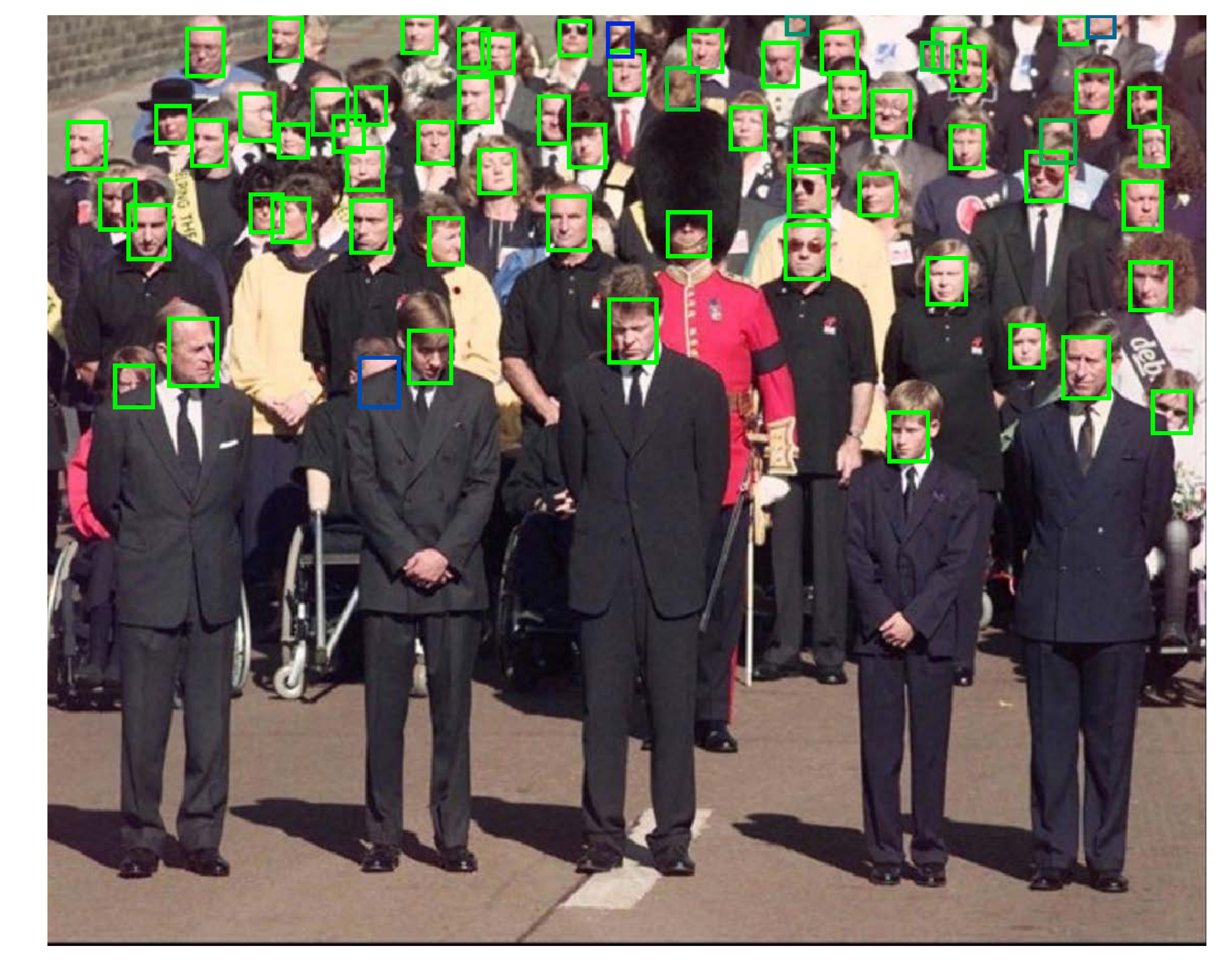}\end{minipage}&
\begin{minipage}{0.21\linewidth}\centering\includegraphics[trim={0cm 0cm 0cm 0cm},clip,width=1\linewidth,height=0.72\linewidth]{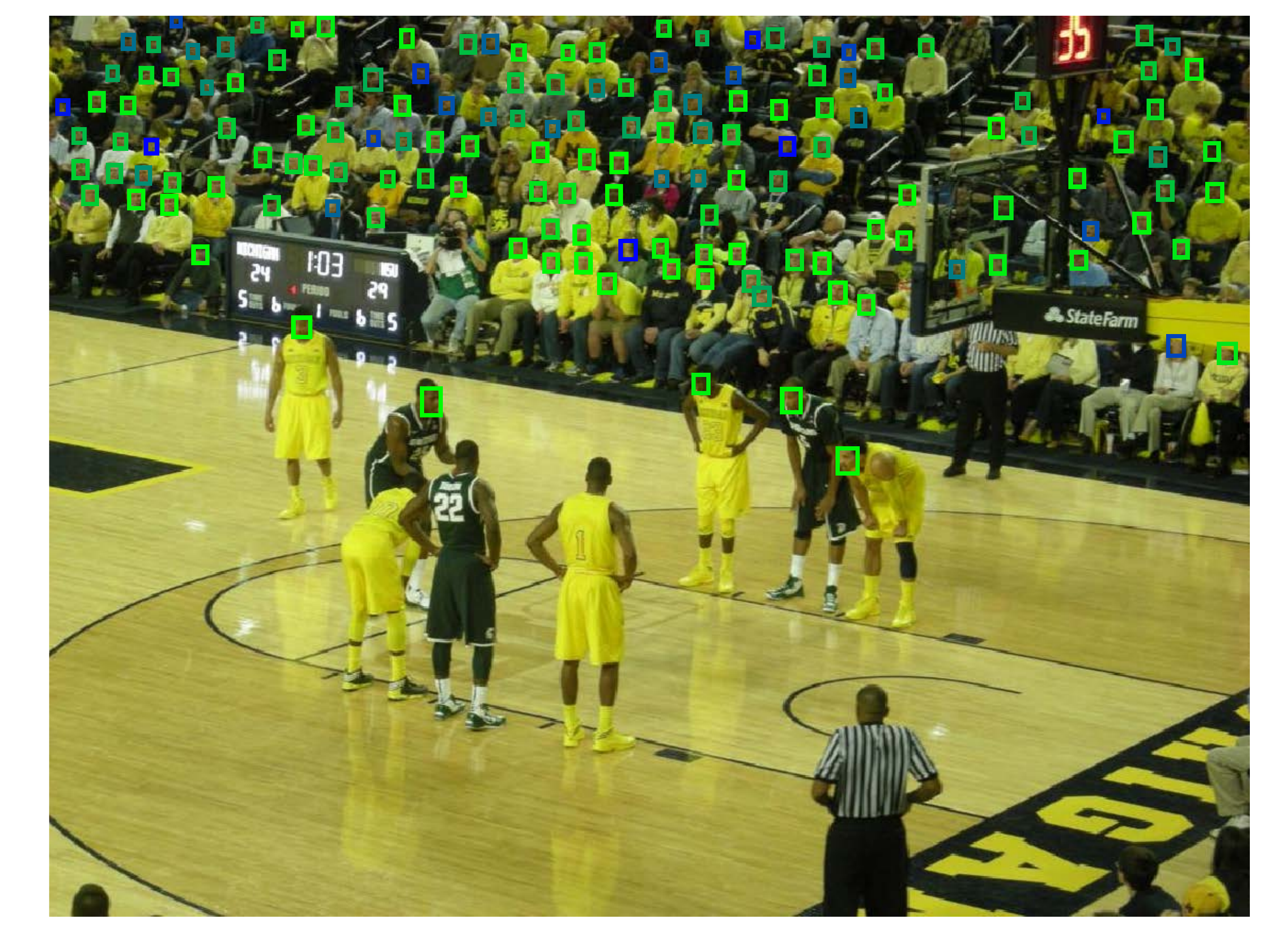}\end{minipage}&
\begin{minipage}{0.21\linewidth}\centering\includegraphics[trim={0cm 0cm 0cm 0cm},clip,width=1\linewidth,height=0.72\linewidth]{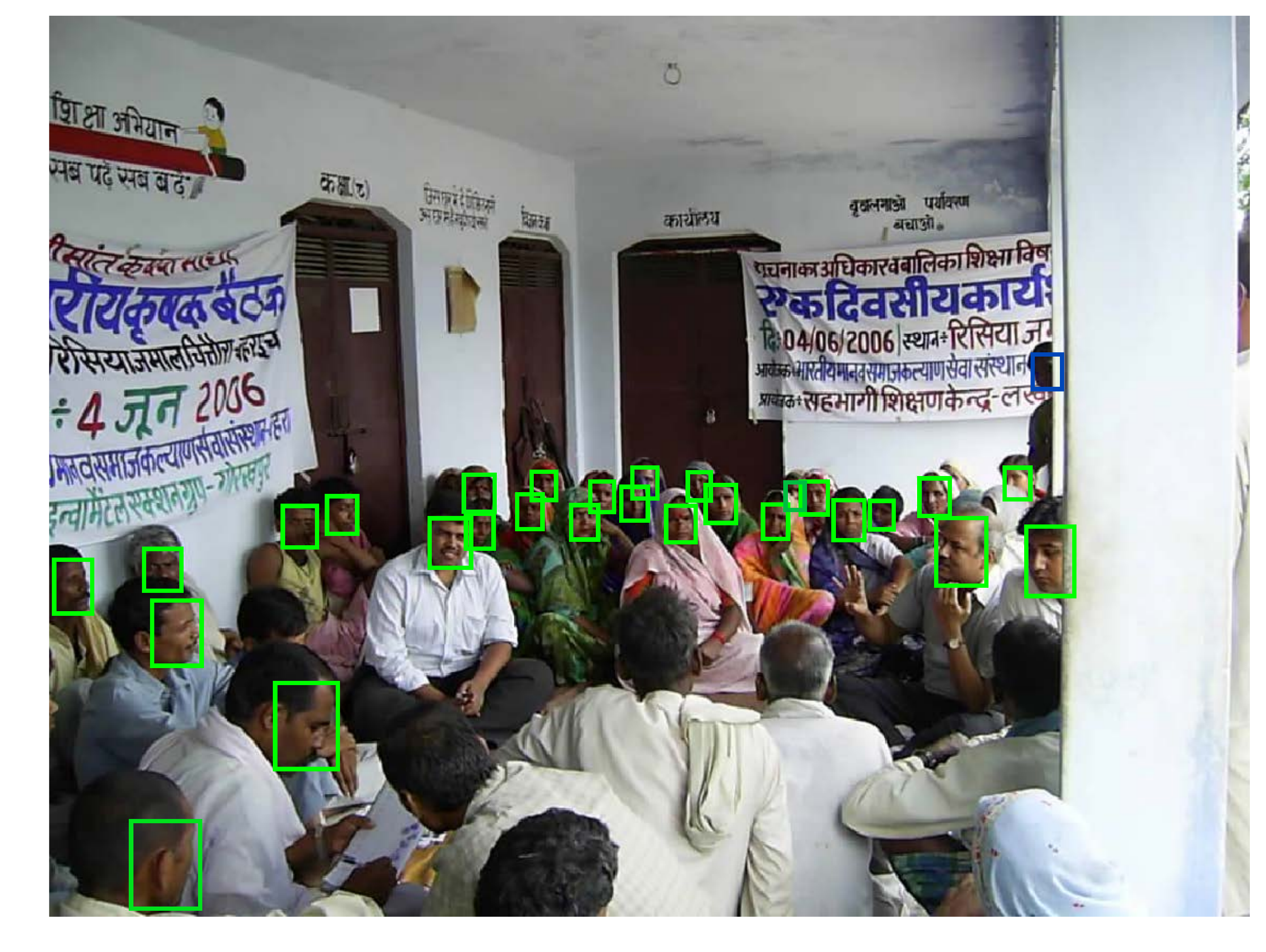}\end{minipage}&
\begin{minipage}{0.21\linewidth}\centering\includegraphics[trim={0cm 0cm 0cm 0cm},clip,width=1\linewidth,height=0.72\linewidth]{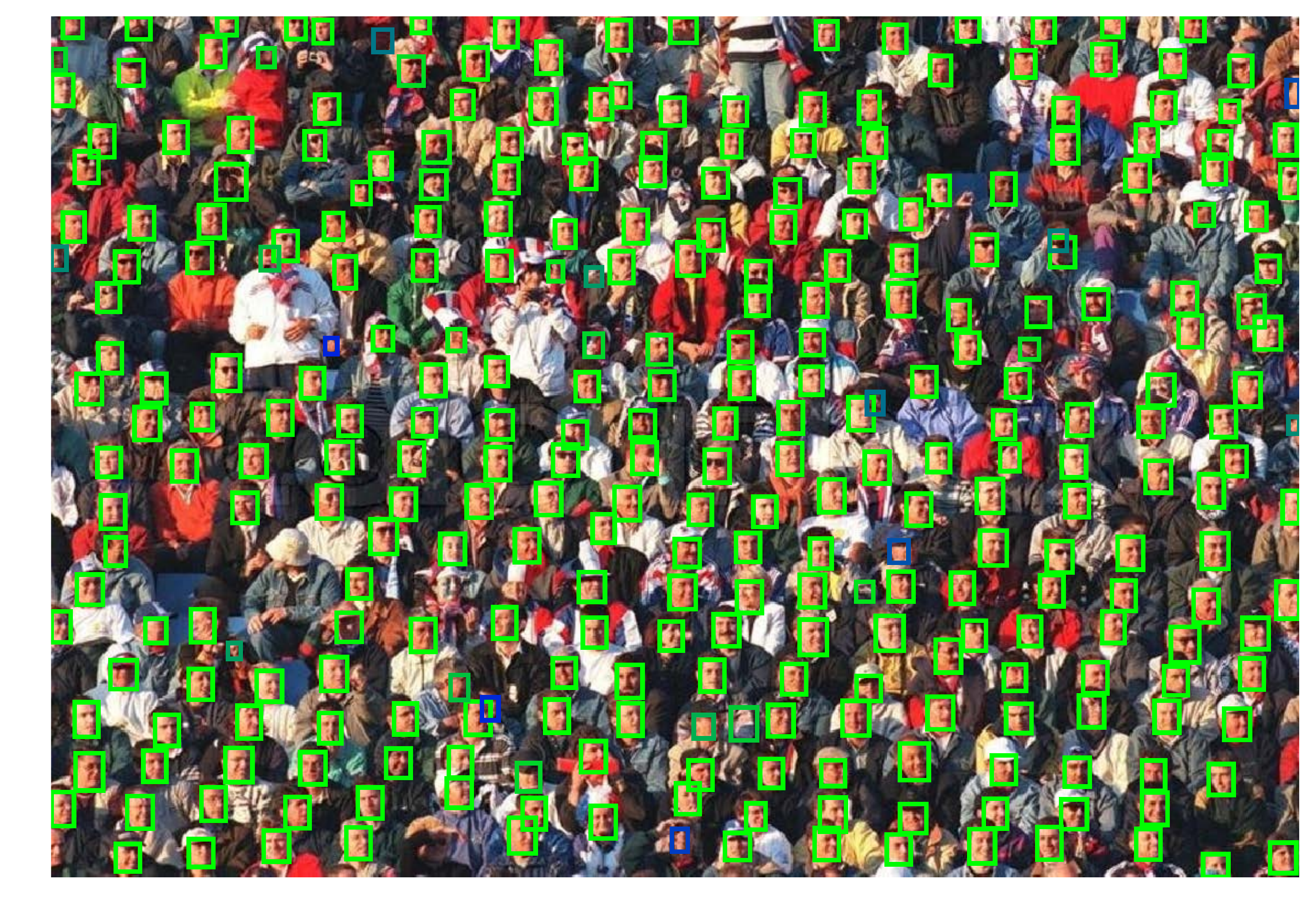}\end{minipage}\\

\hline
\begin{minipage}{0.21\linewidth}\centering\includegraphics[trim={0cm 0cm 0cm 0cm},clip,width=1\linewidth,height=0.72\linewidth]{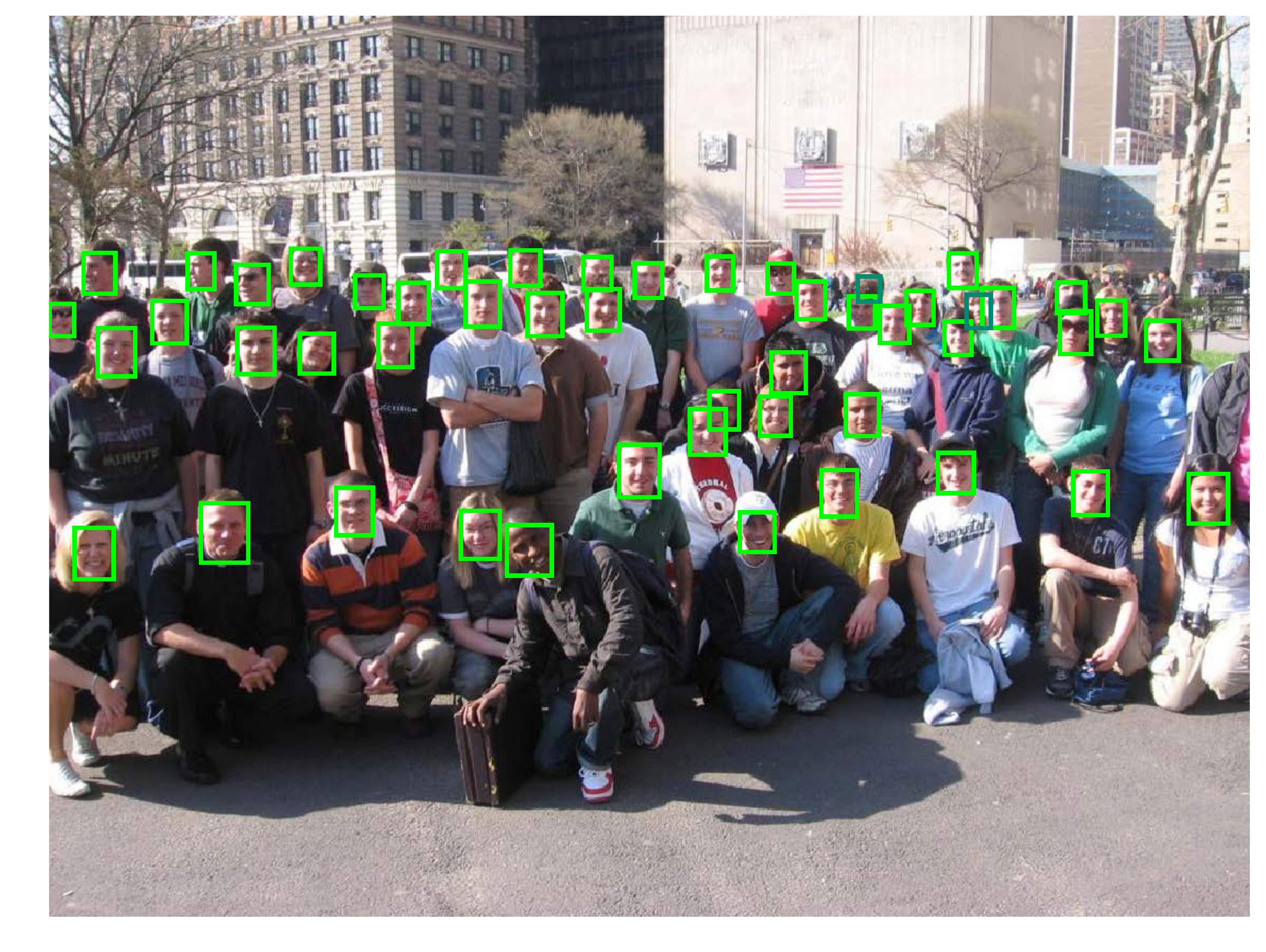}\end{minipage}&
\begin{minipage}{0.21\linewidth}\centering\includegraphics[trim={0cm 0cm 0cm 0cm},clip,width=1\linewidth,height=0.72\linewidth]{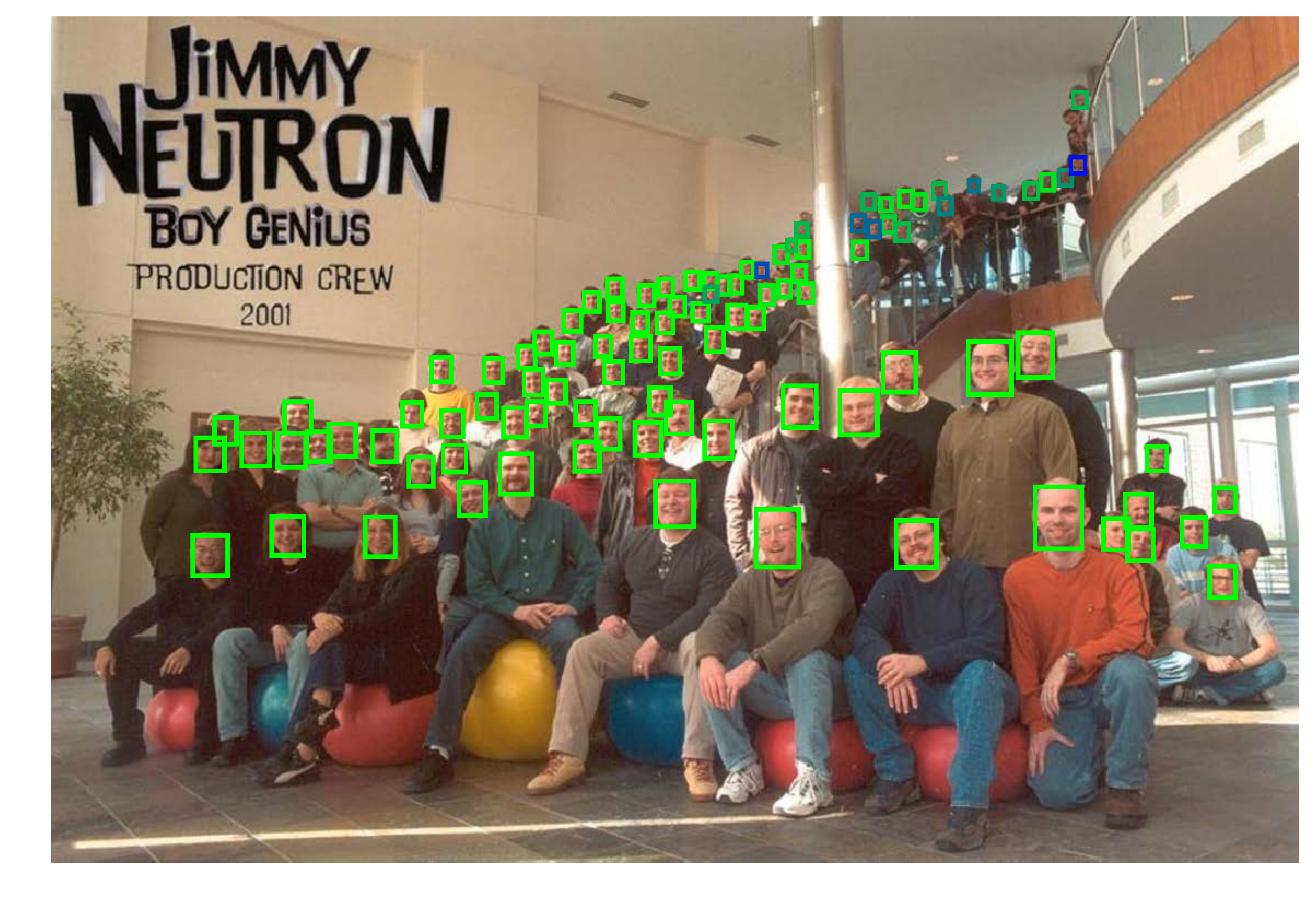}\end{minipage}&
\begin{minipage}{0.21\linewidth}\centering\includegraphics[trim={0cm 0cm 0cm 0cm},clip,width=1\linewidth,height=0.72\linewidth]{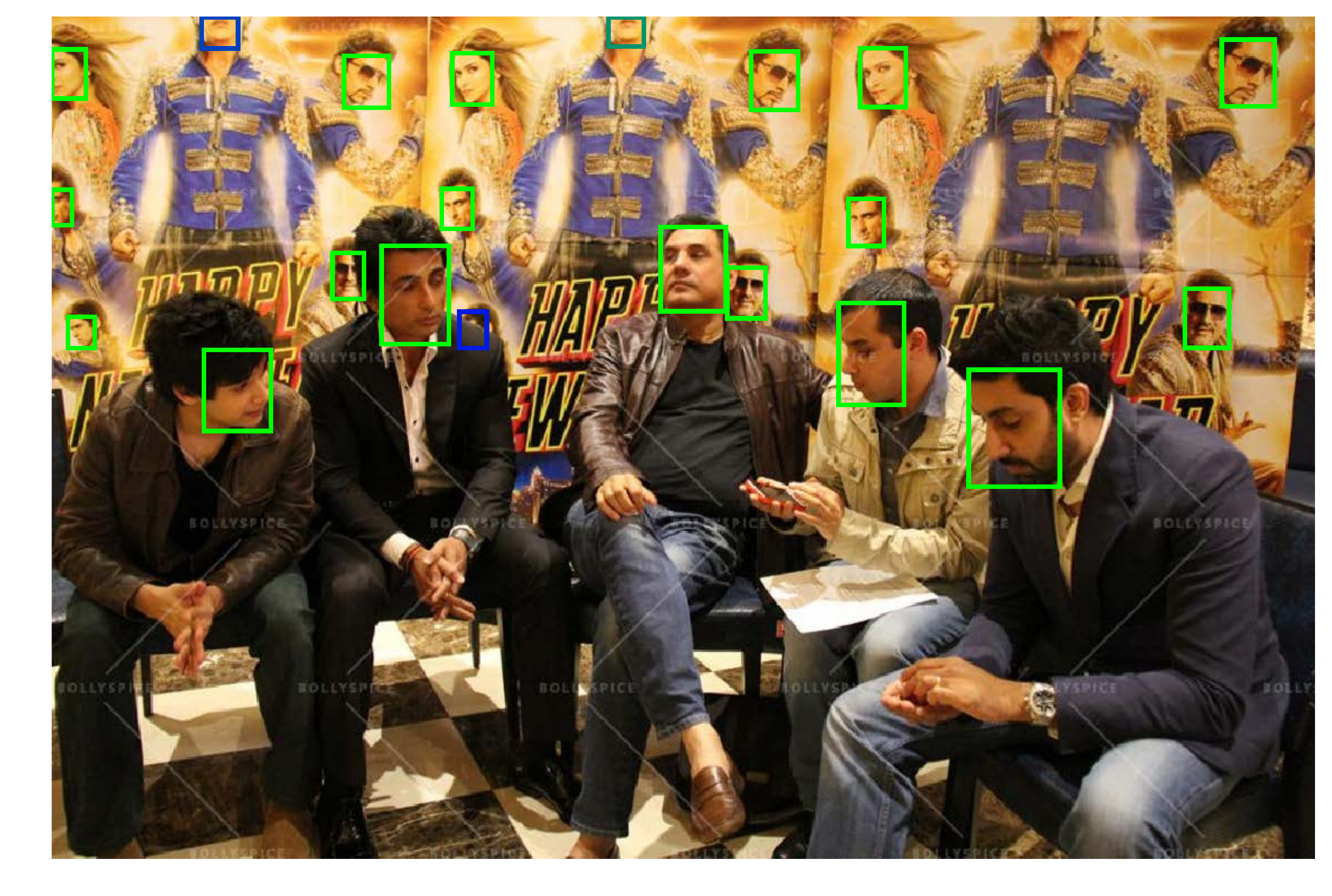}\end{minipage}&
\begin{minipage}{0.21\linewidth}\centering\includegraphics[trim={0cm 0cm 0cm 0cm},clip,width=1\linewidth,height=0.72\linewidth]{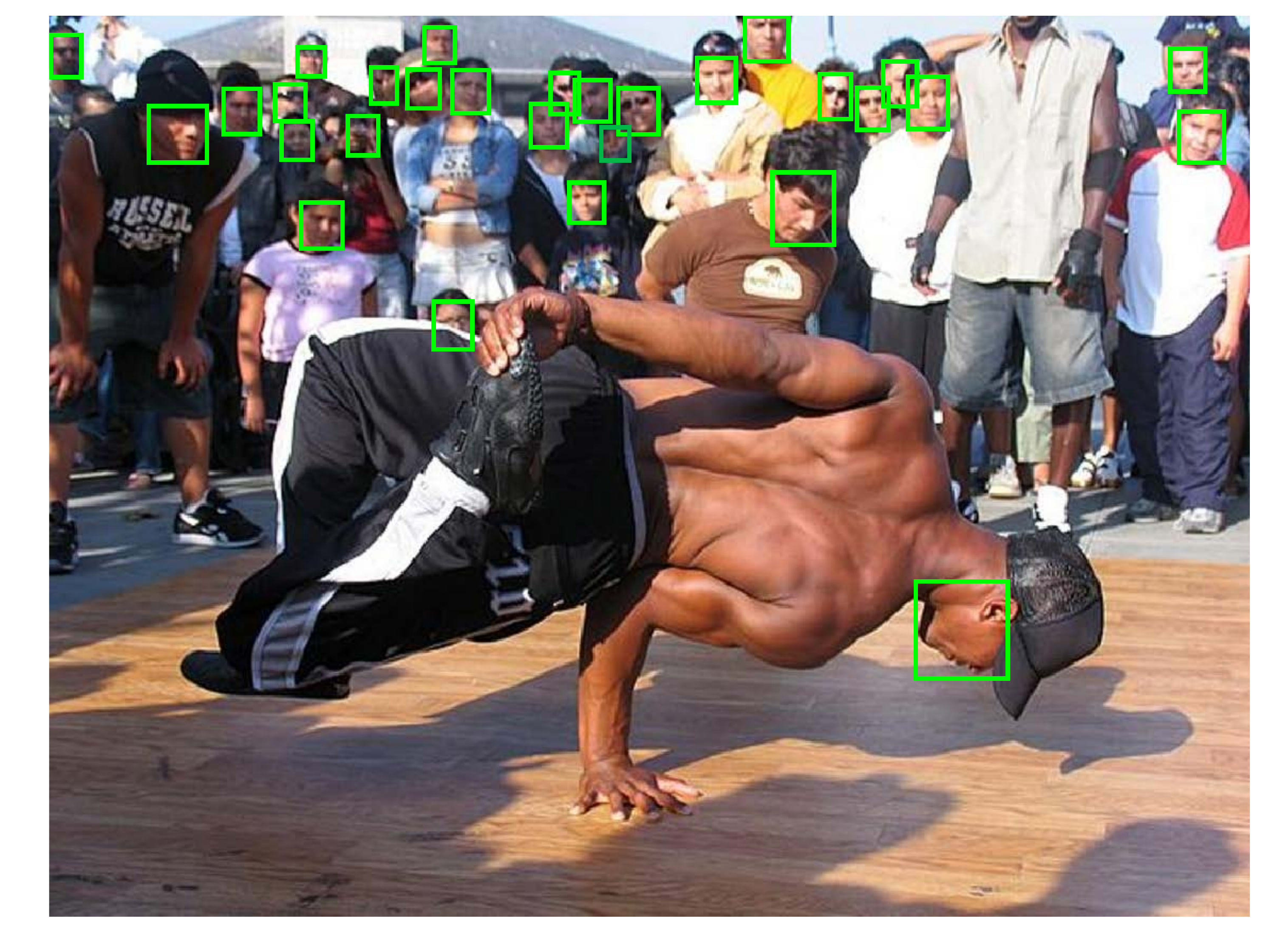}\end{minipage}\\

\hline
\begin{minipage}{0.21\linewidth}\centering\includegraphics[trim={0cm 0cm 0cm 0cm},clip,width=1\linewidth,height=0.72\linewidth]{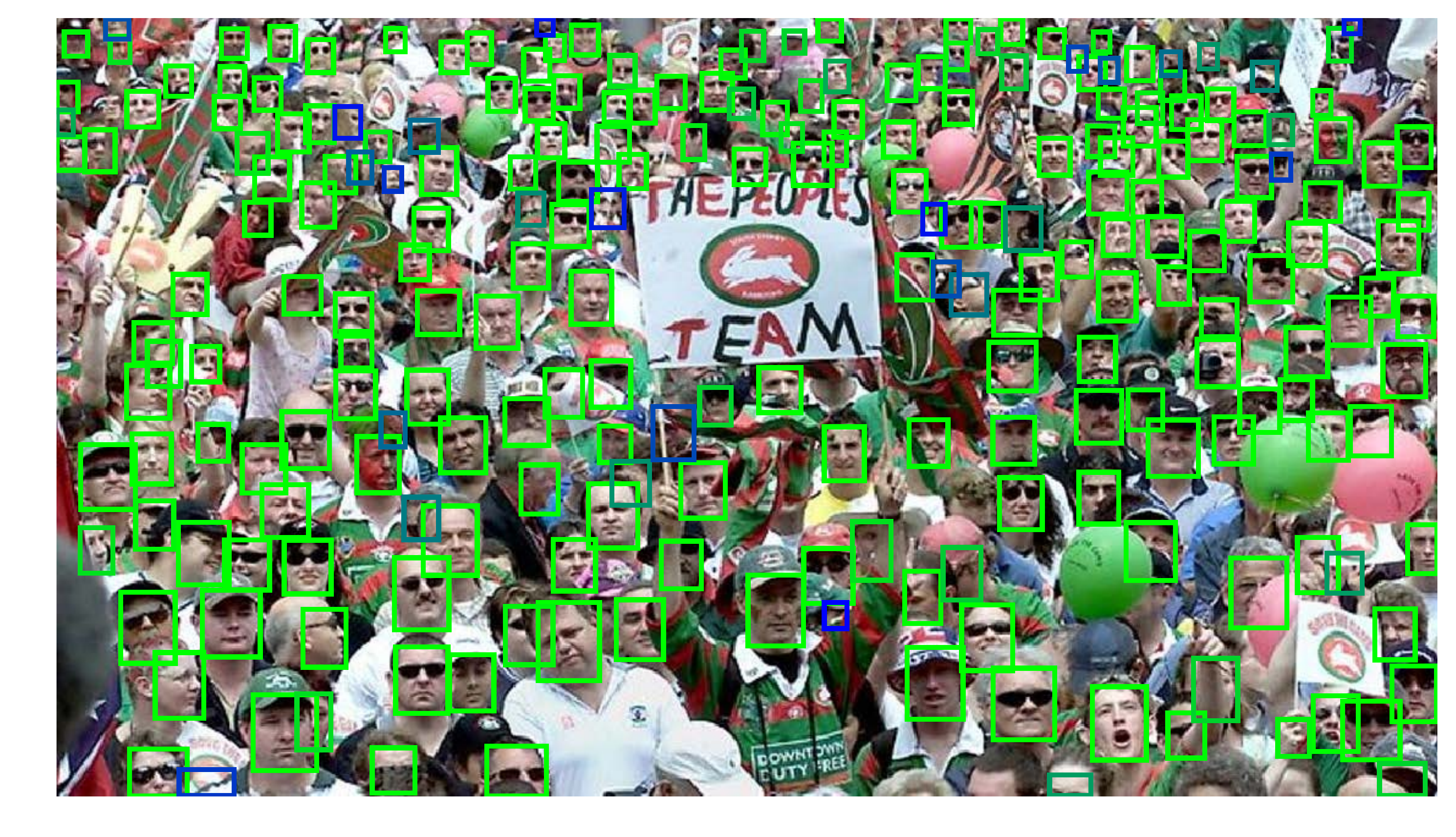}\end{minipage}&
\begin{minipage}{0.21\linewidth}\centering\includegraphics[trim={0cm 0cm 0cm 0cm},clip,width=1\linewidth,height=0.72\linewidth]{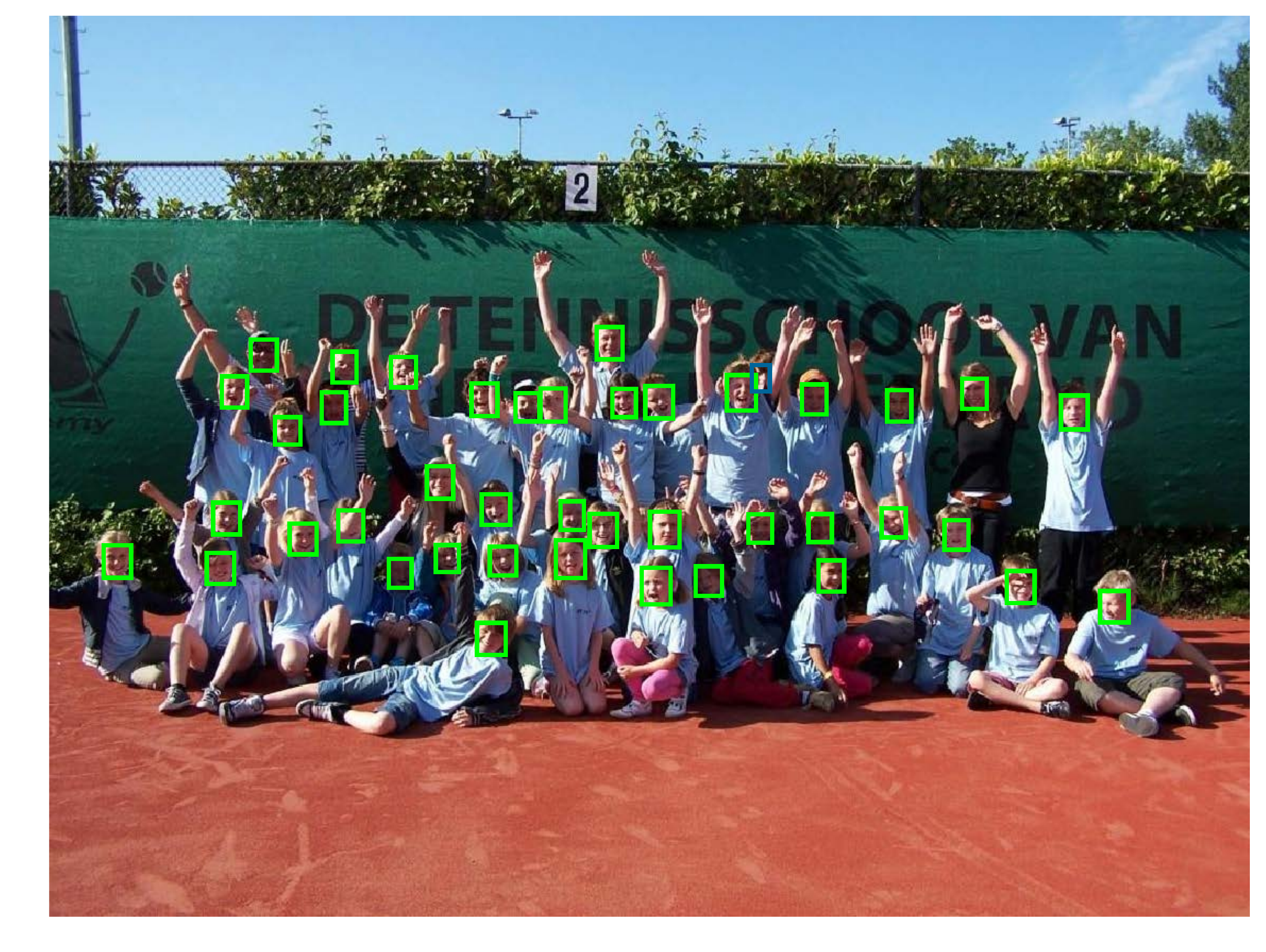}\end{minipage}&
\begin{minipage}{0.21\linewidth}\centering\includegraphics[trim={0cm 0cm 0cm 0cm},clip,width=1\linewidth,height=0.72\linewidth]{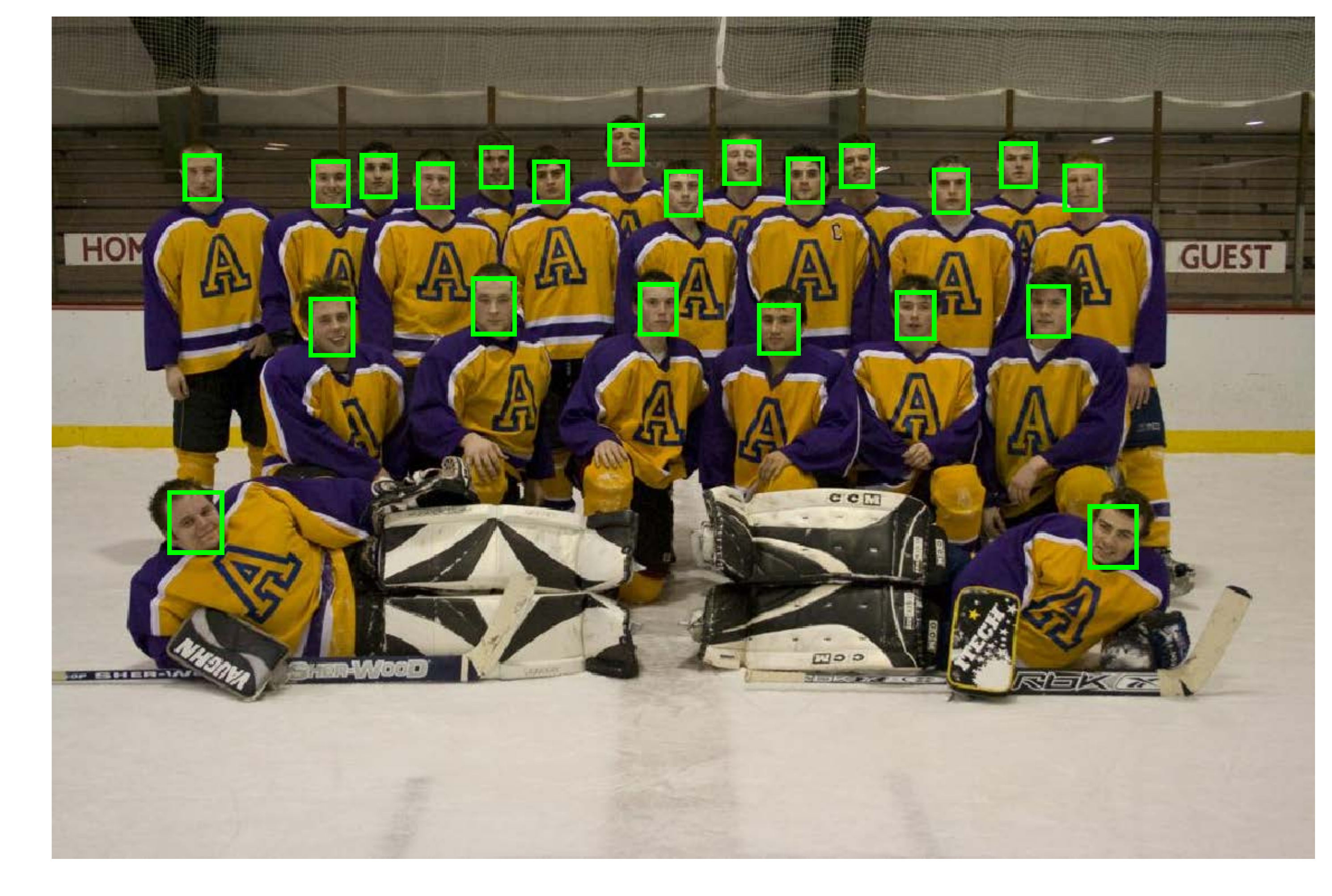}\end{minipage}&
\begin{minipage}{0.21\linewidth}\centering\includegraphics[trim={0cm 0cm 0cm 0cm},clip,width=1\linewidth,height=0.72\linewidth]{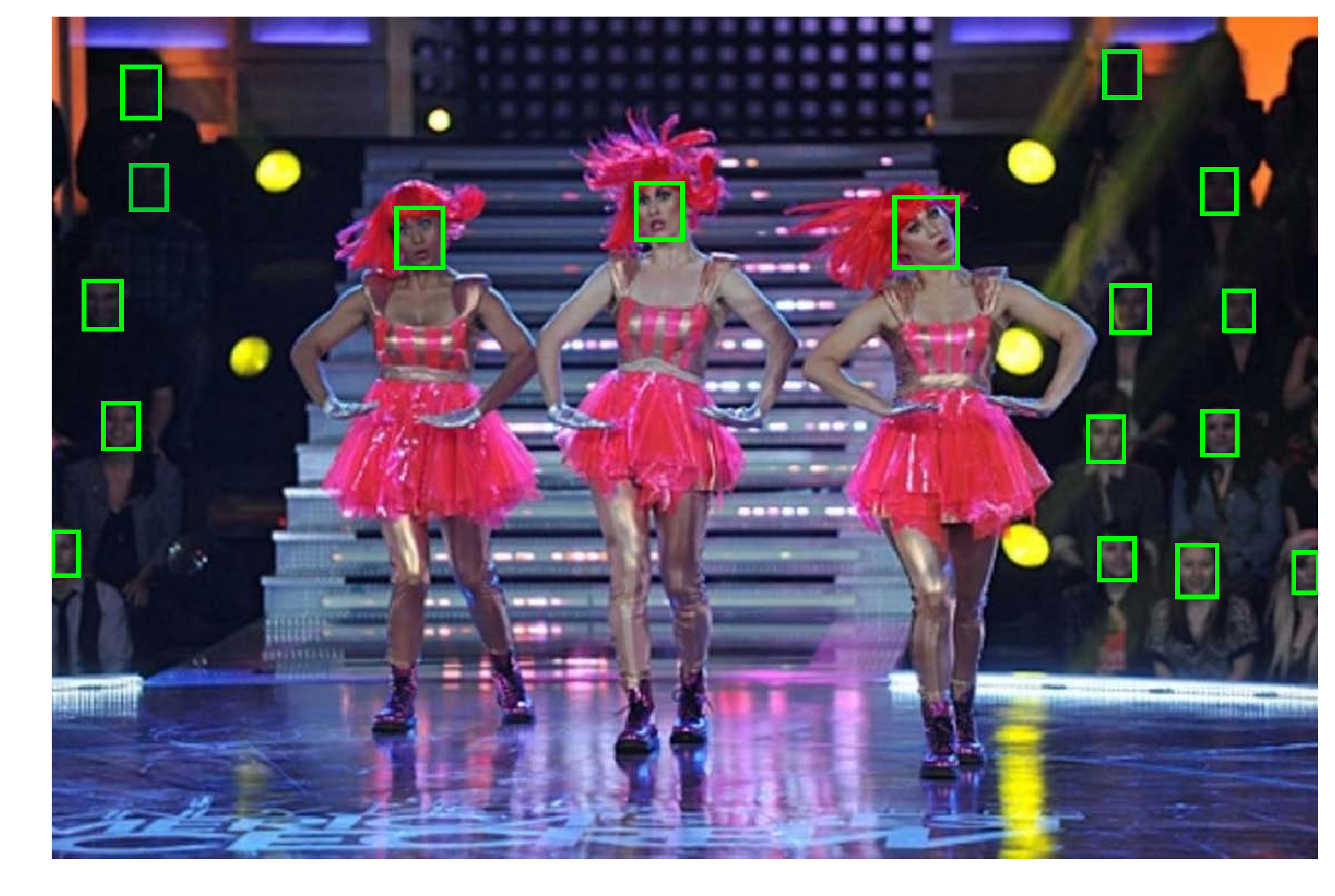}\end{minipage}\\

\hline
\begin{minipage}{0.21\linewidth}\centering\includegraphics[trim={0cm 0cm 0cm 0cm},clip,width=1\linewidth,height=0.72\linewidth]{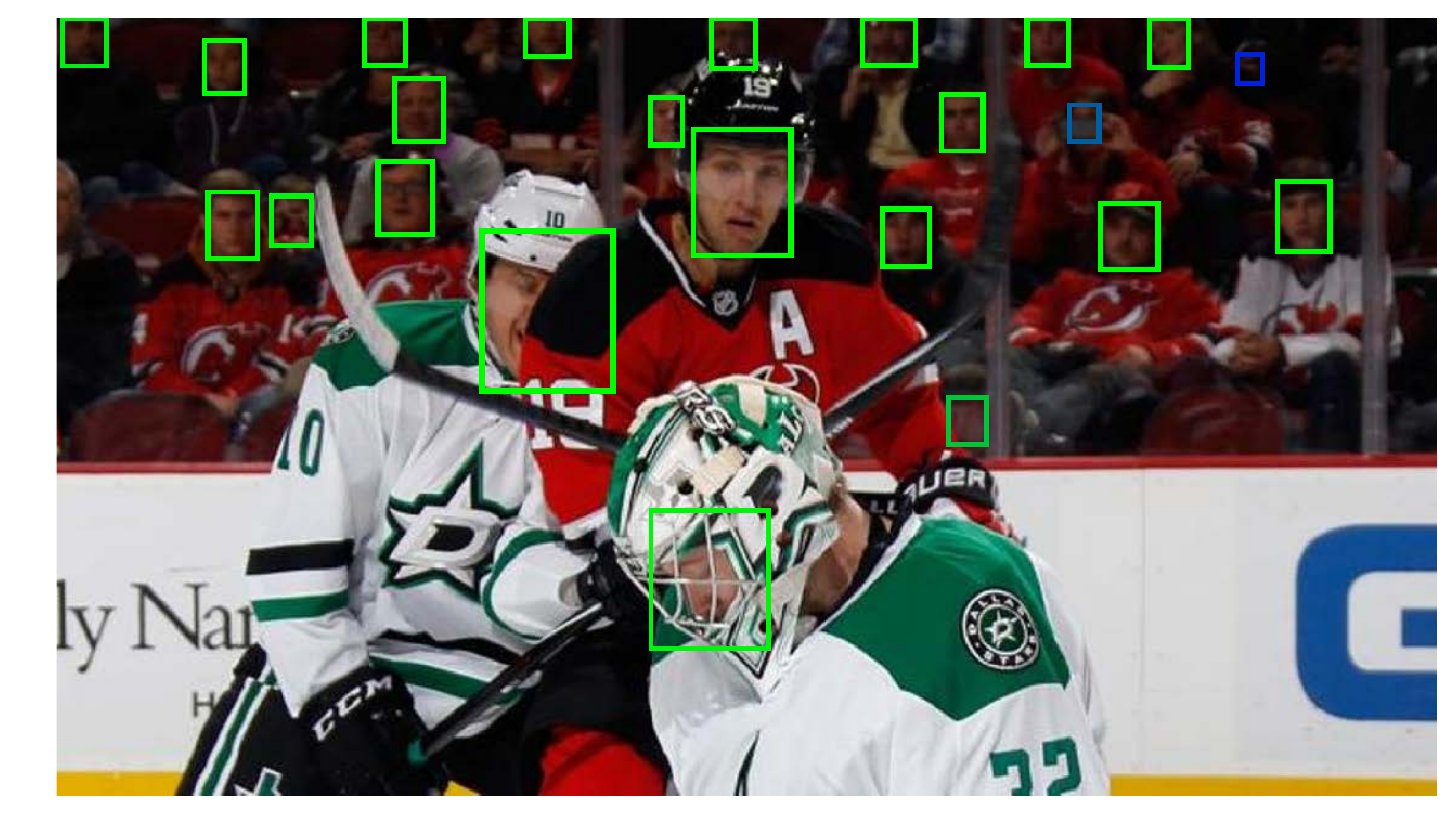}\end{minipage}&
\begin{minipage}{0.21\linewidth}\centering\includegraphics[trim={0cm 0cm 0cm 0cm},clip,width=1\linewidth,height=0.72\linewidth]{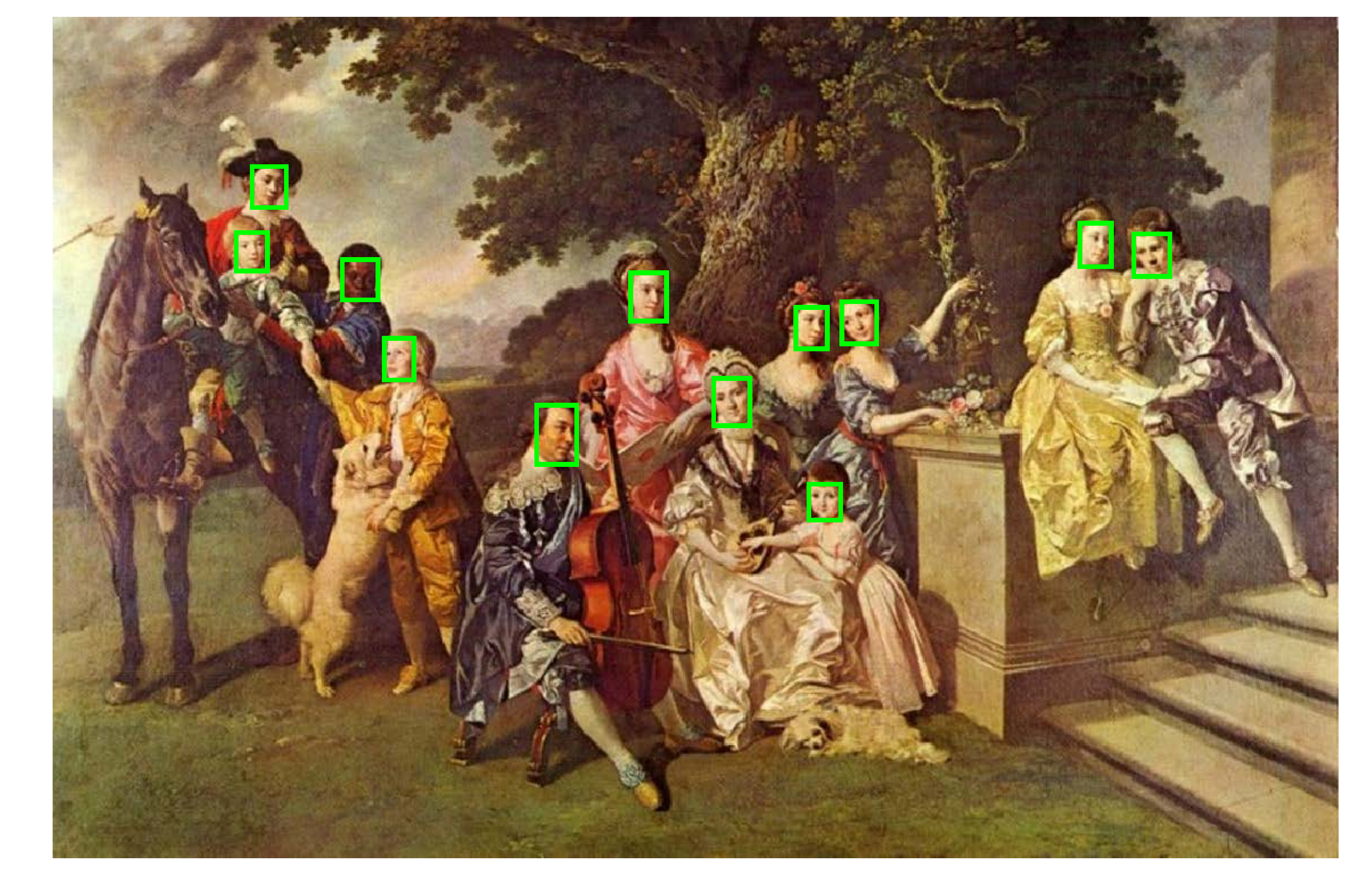}\end{minipage}&
\begin{minipage}{0.21\linewidth}\centering\includegraphics[trim={0cm 0cm 0cm 0cm},clip,width=1\linewidth,height=0.72\linewidth]{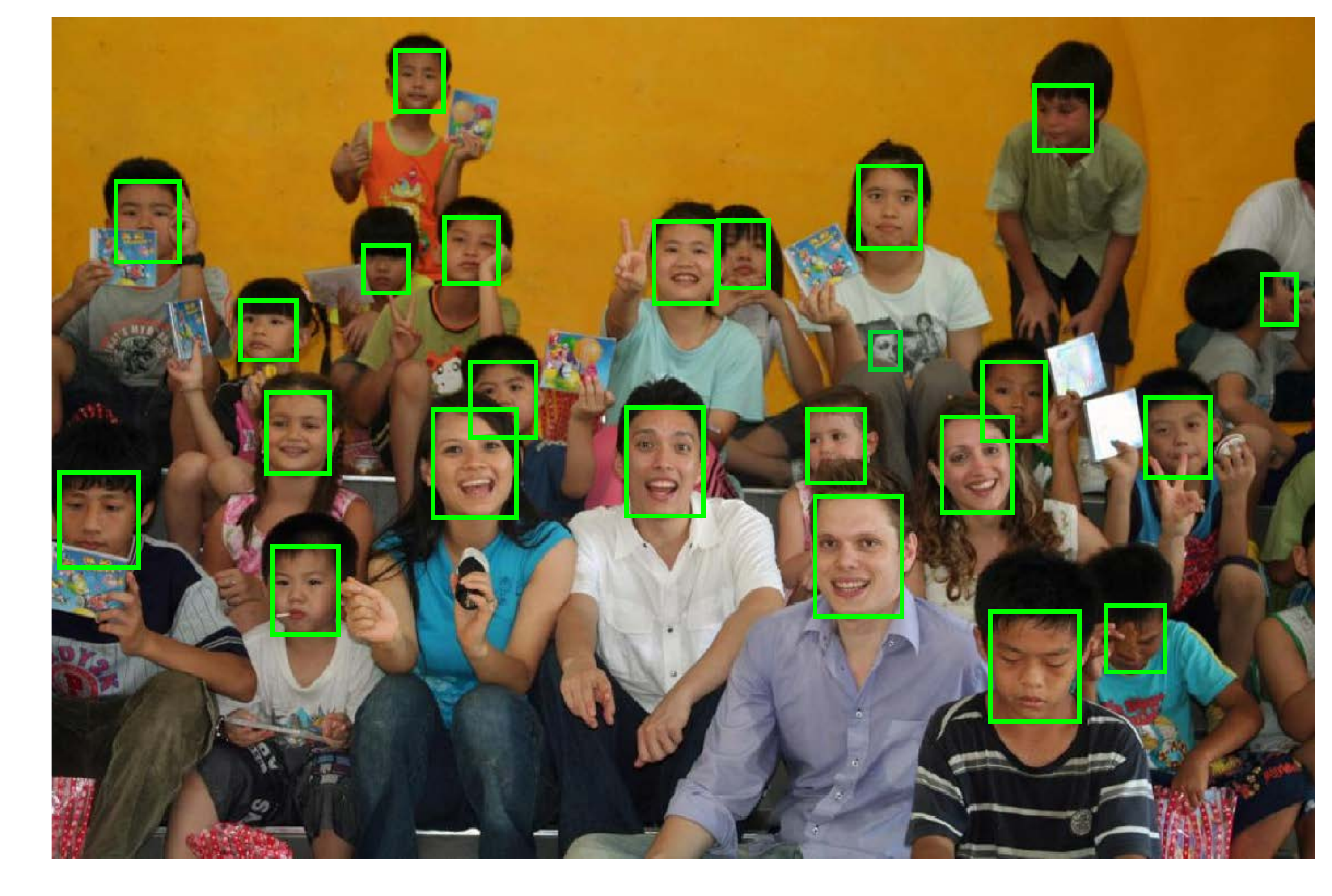}\end{minipage}&
\begin{minipage}{0.21\linewidth}\centering\includegraphics[trim={0cm 0cm 0cm 0cm},clip,width=1\linewidth,height=0.72\linewidth]{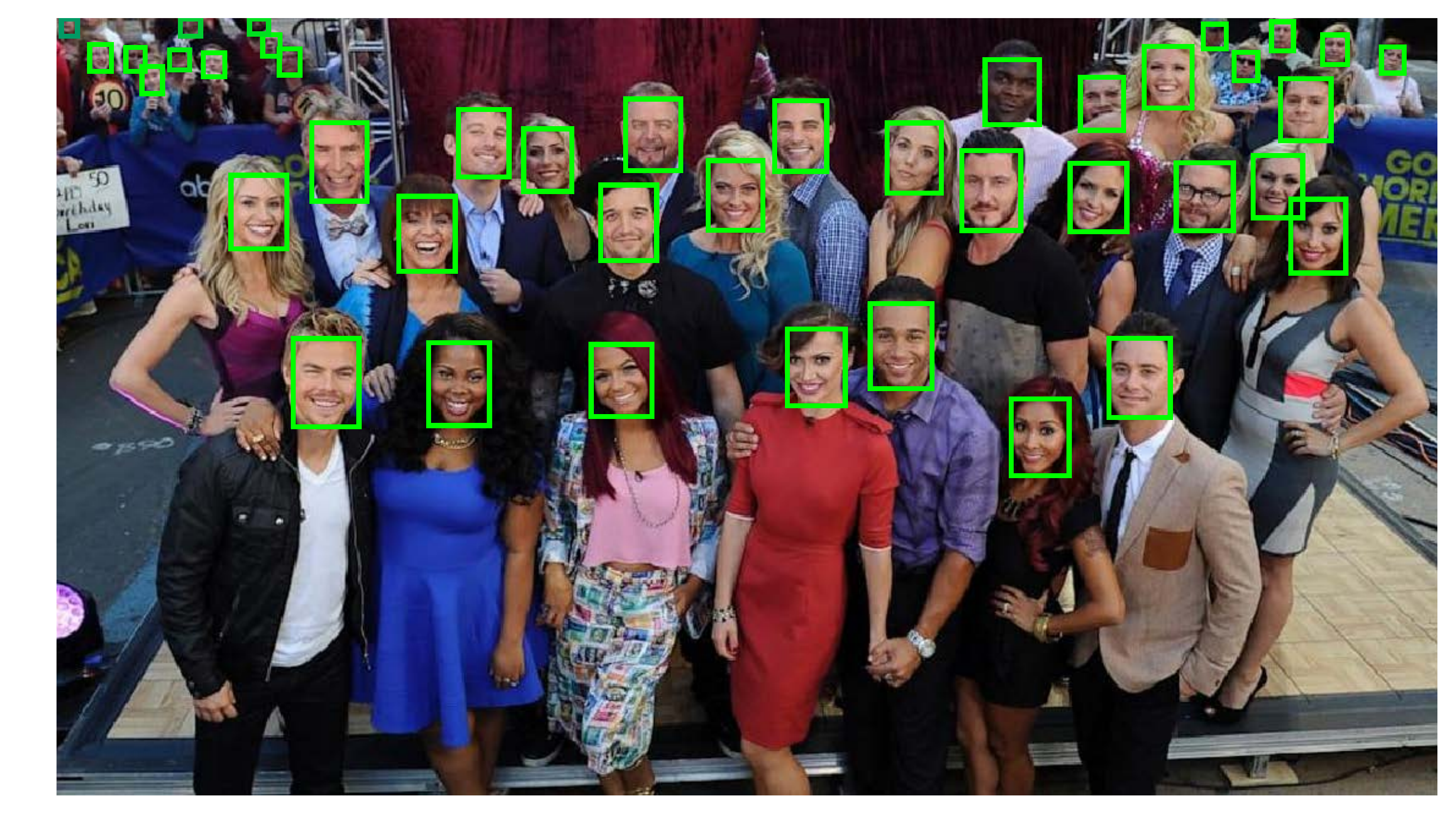}\end{minipage}\\

\hline
\begin{minipage}{0.21\linewidth}\centering\includegraphics[trim={0cm 0cm 0cm 0cm},clip,width=1\linewidth,height=0.72\linewidth]{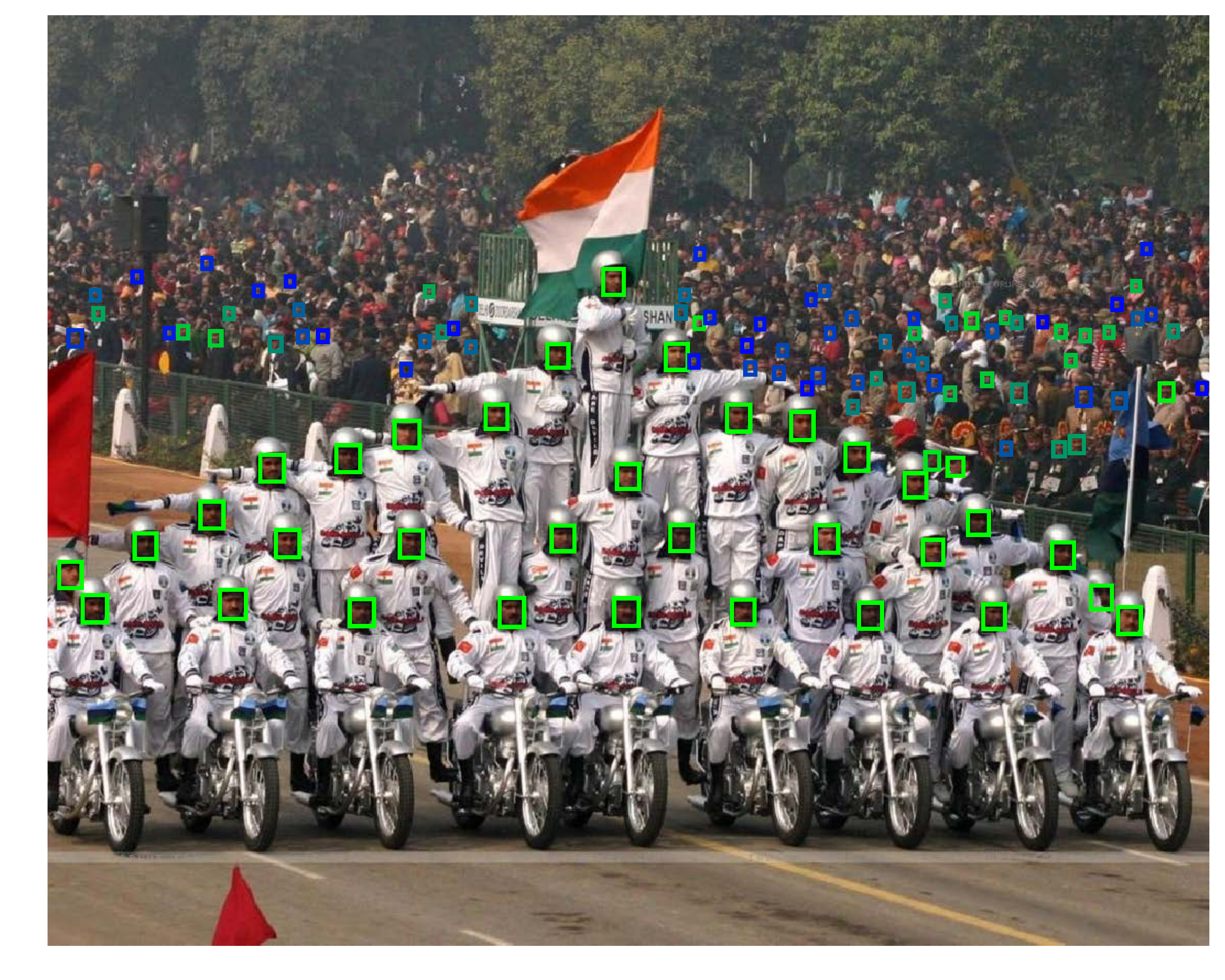}\end{minipage}&
\begin{minipage}{0.21\linewidth}\centering\includegraphics[trim={0cm 0cm 0cm 0cm},clip,width=1\linewidth,height=0.72\linewidth]{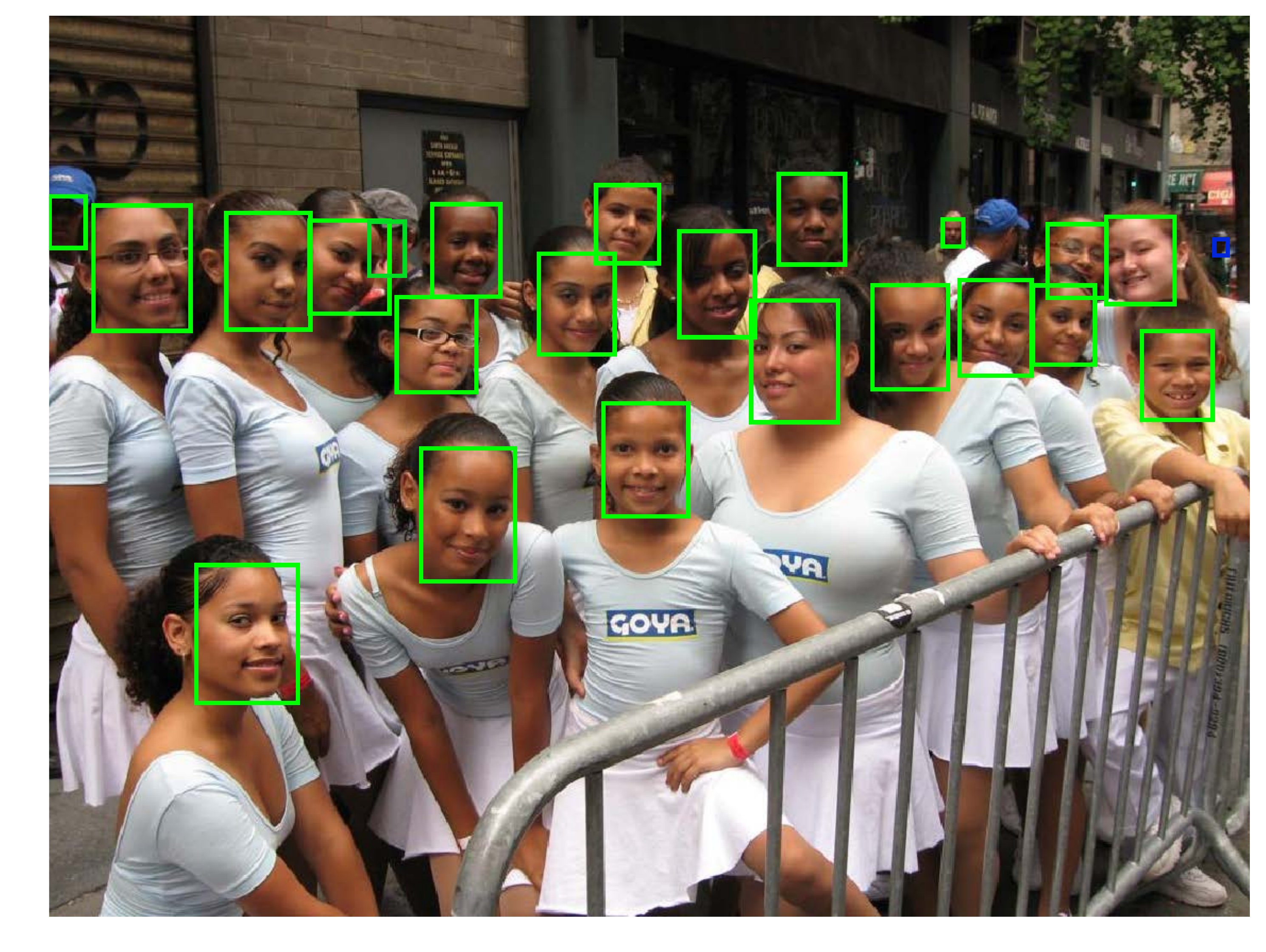}\end{minipage}&
\begin{minipage}{0.21\linewidth}\centering\includegraphics[trim={0cm 0cm 0cm 0cm},clip,width=1\linewidth,height=0.72\linewidth]{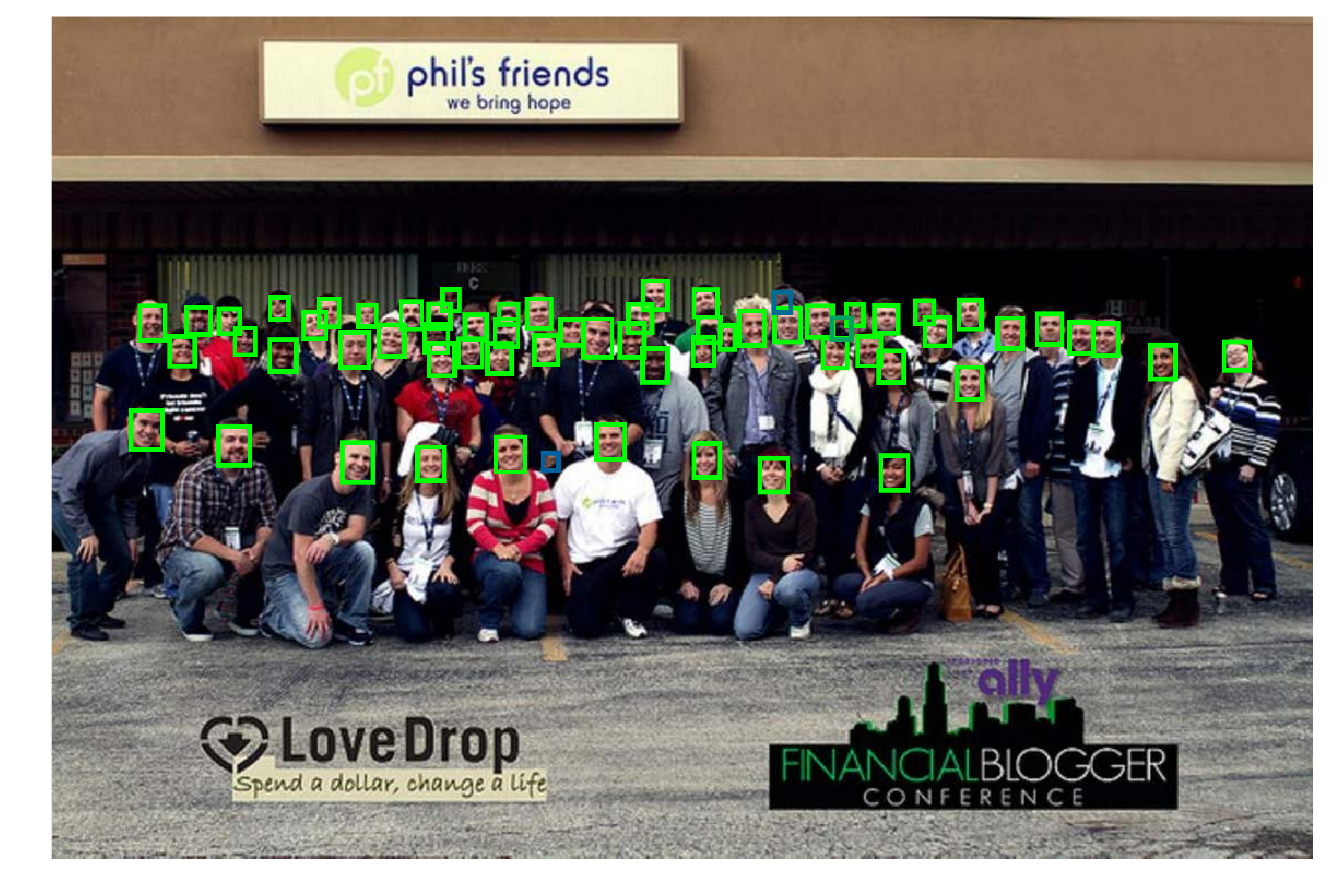}\end{minipage}&
\begin{minipage}{0.21\linewidth}\centering\includegraphics[trim={0cm 0cm 0cm 0cm},clip,width=1\linewidth,height=0.72\linewidth]{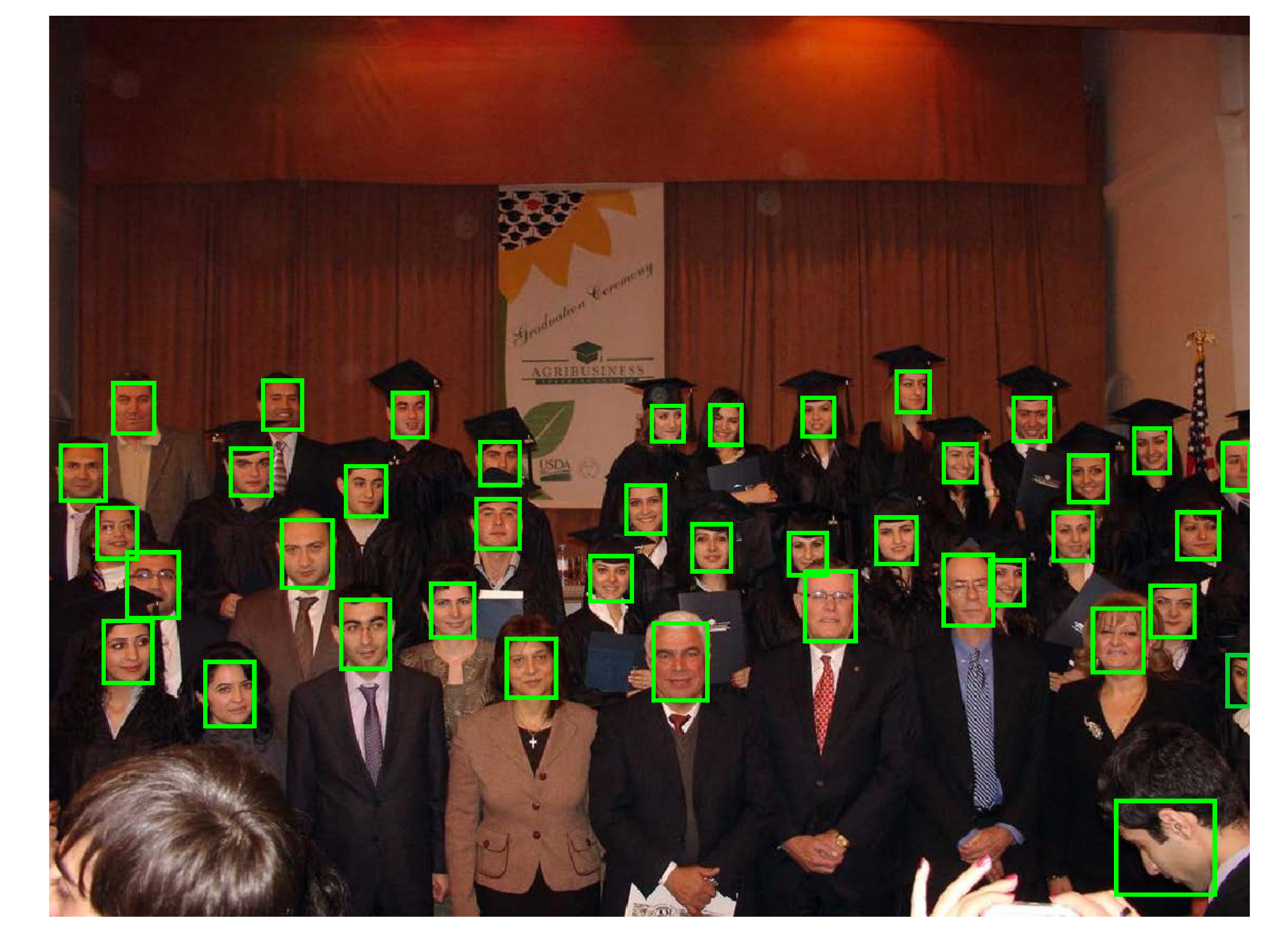}\end{minipage}\\

\hline
\begin{minipage}{0.21\linewidth}\centering\includegraphics[trim={0cm 0cm 0cm 0cm},clip,width=1\linewidth,height=0.72\linewidth]{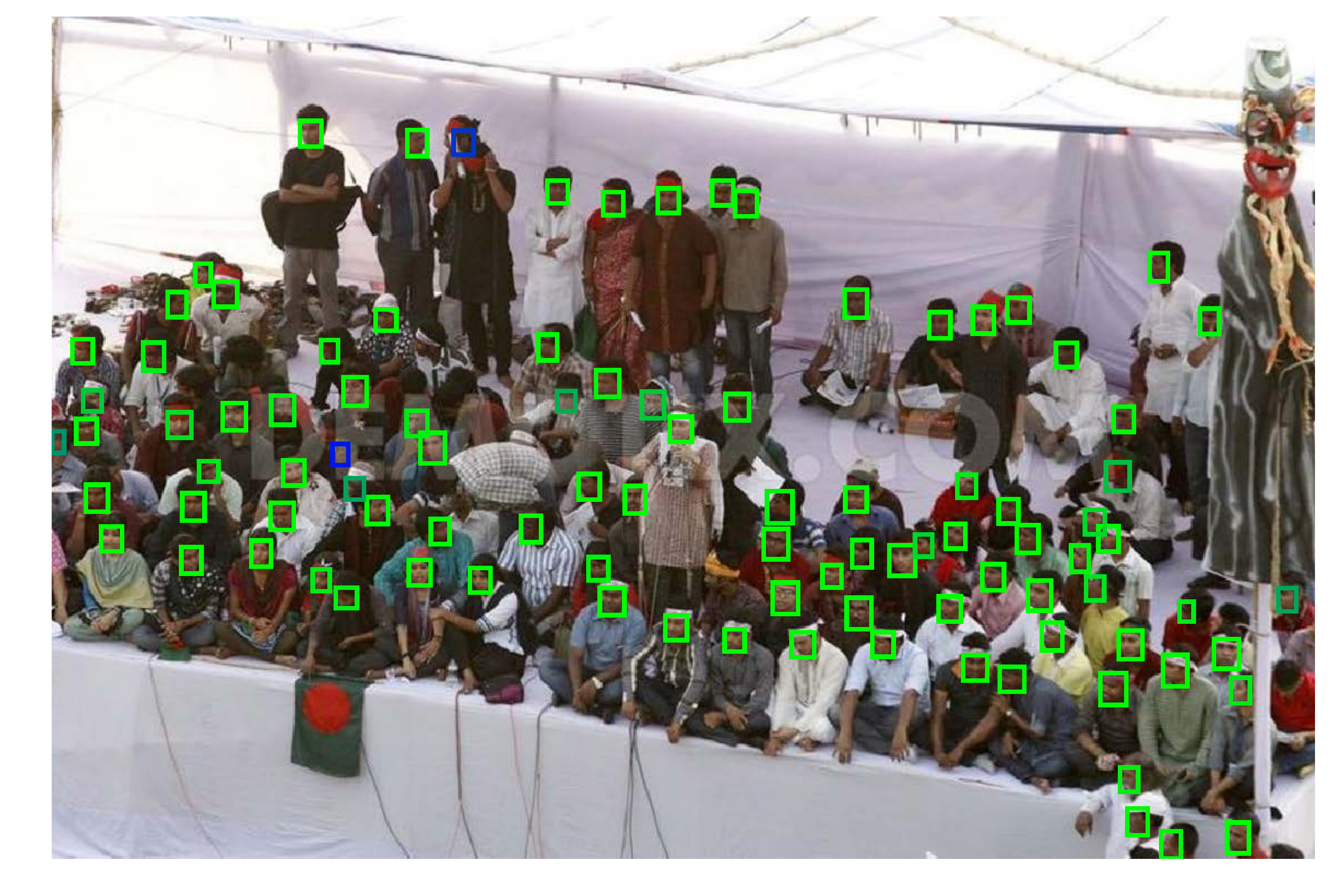}\end{minipage}&
\begin{minipage}{0.21\linewidth}\centering\includegraphics[trim={0cm 0cm 0cm 0cm},clip,width=1\linewidth,height=0.72\linewidth]{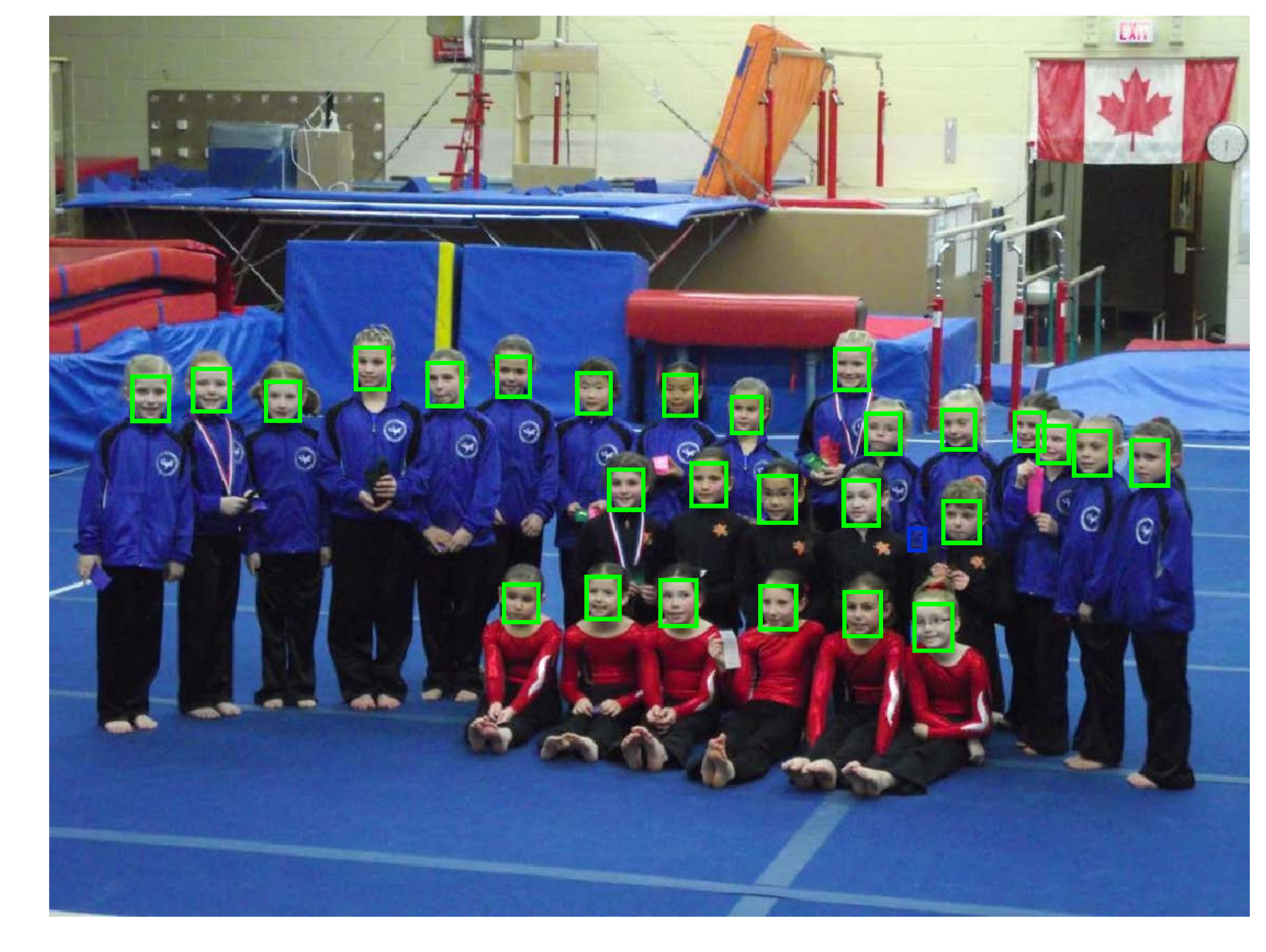}\end{minipage}&
\begin{minipage}{0.21\linewidth}\centering\includegraphics[trim={0cm 0cm 0cm 0cm},clip,width=1\linewidth,height=0.72\linewidth]{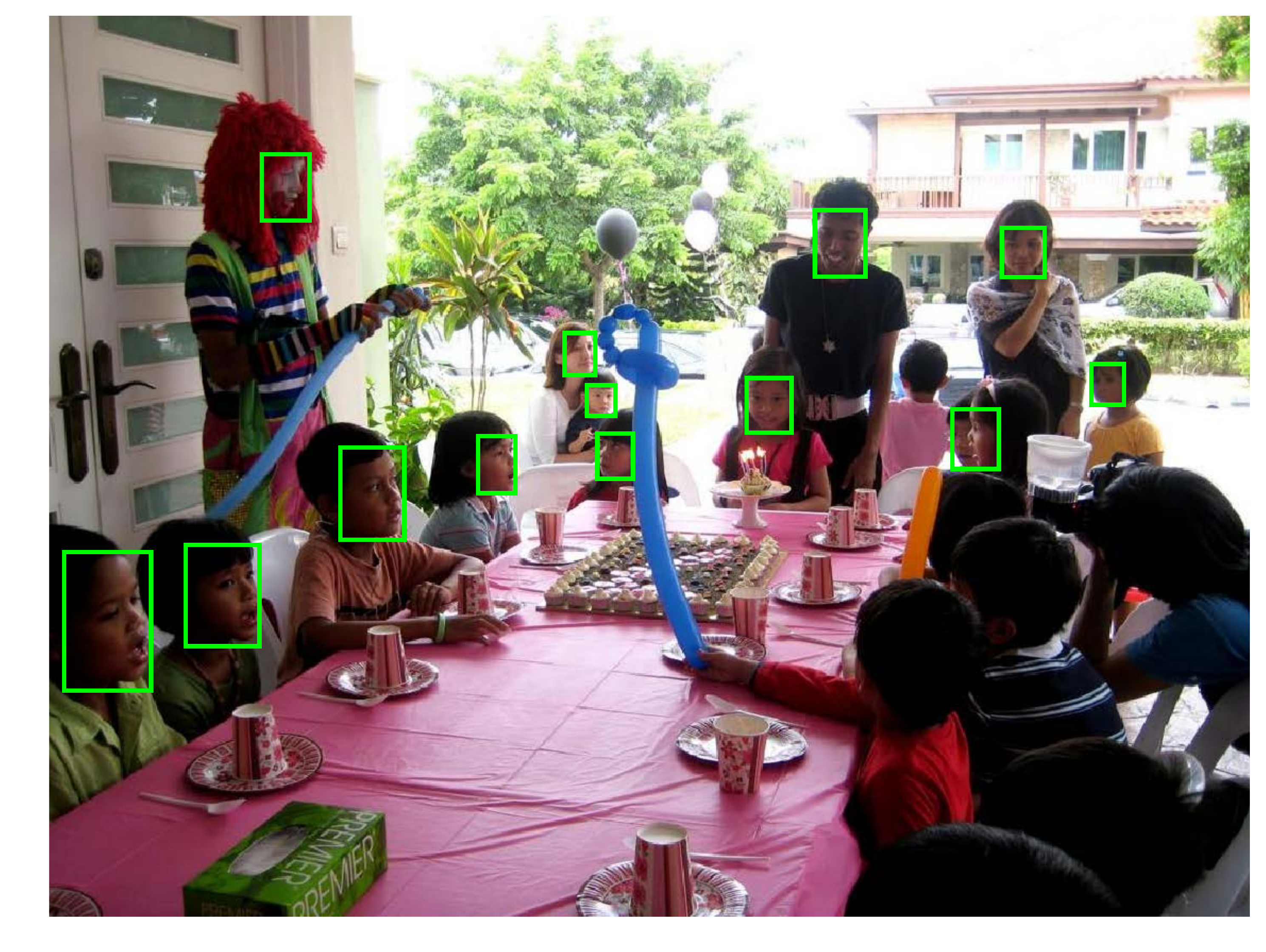}\end{minipage}&
\begin{minipage}{0.21\linewidth}\centering\includegraphics[trim={0cm 0cm 0cm 0cm},clip,width=1\linewidth,height=0.72\linewidth]{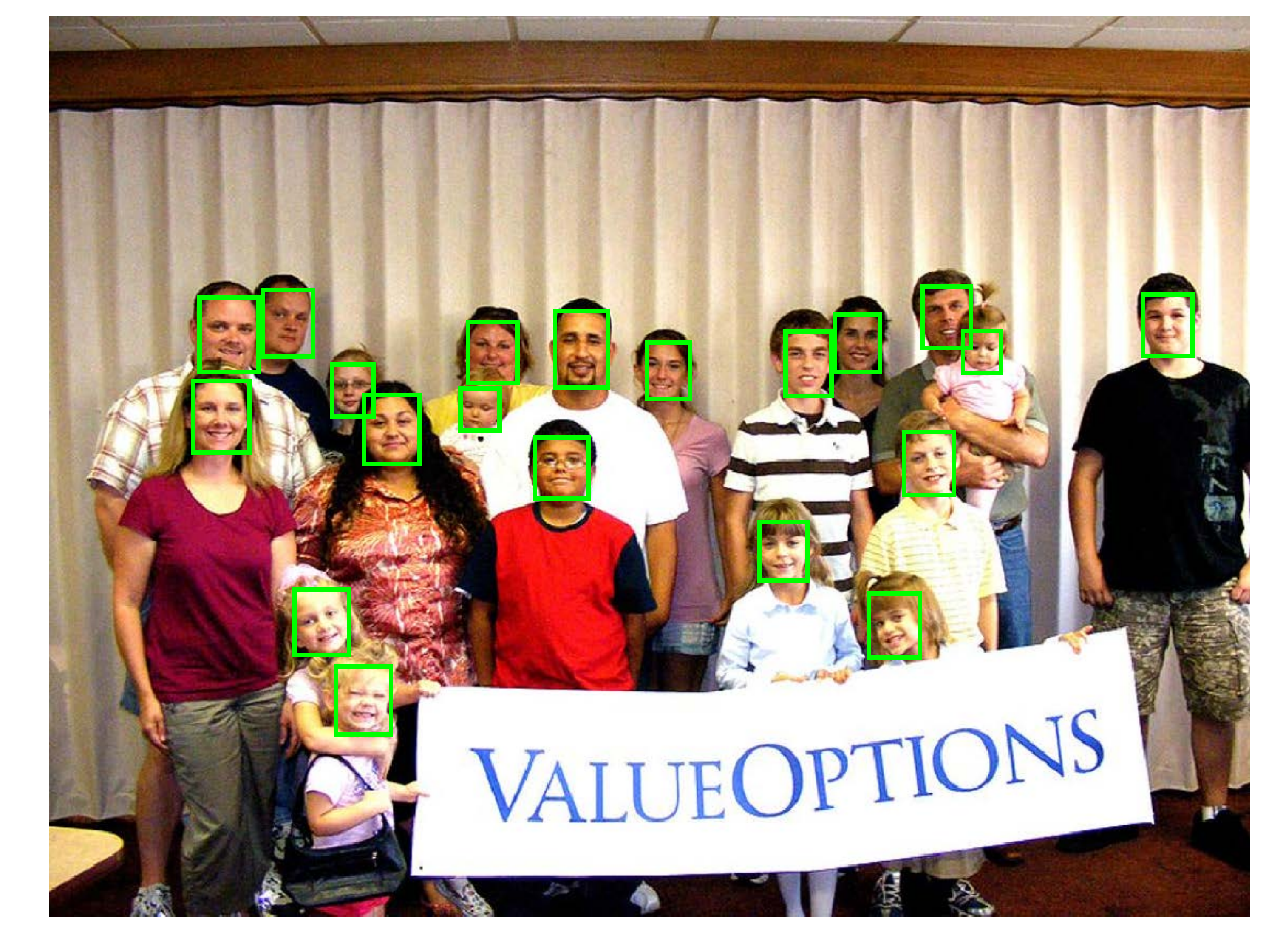}\end{minipage}\\

\hline
\begin{minipage}{0.21\linewidth}\centering\includegraphics[trim={0cm 0cm 0cm 0cm},clip,width=1\linewidth,height=0.72\linewidth]{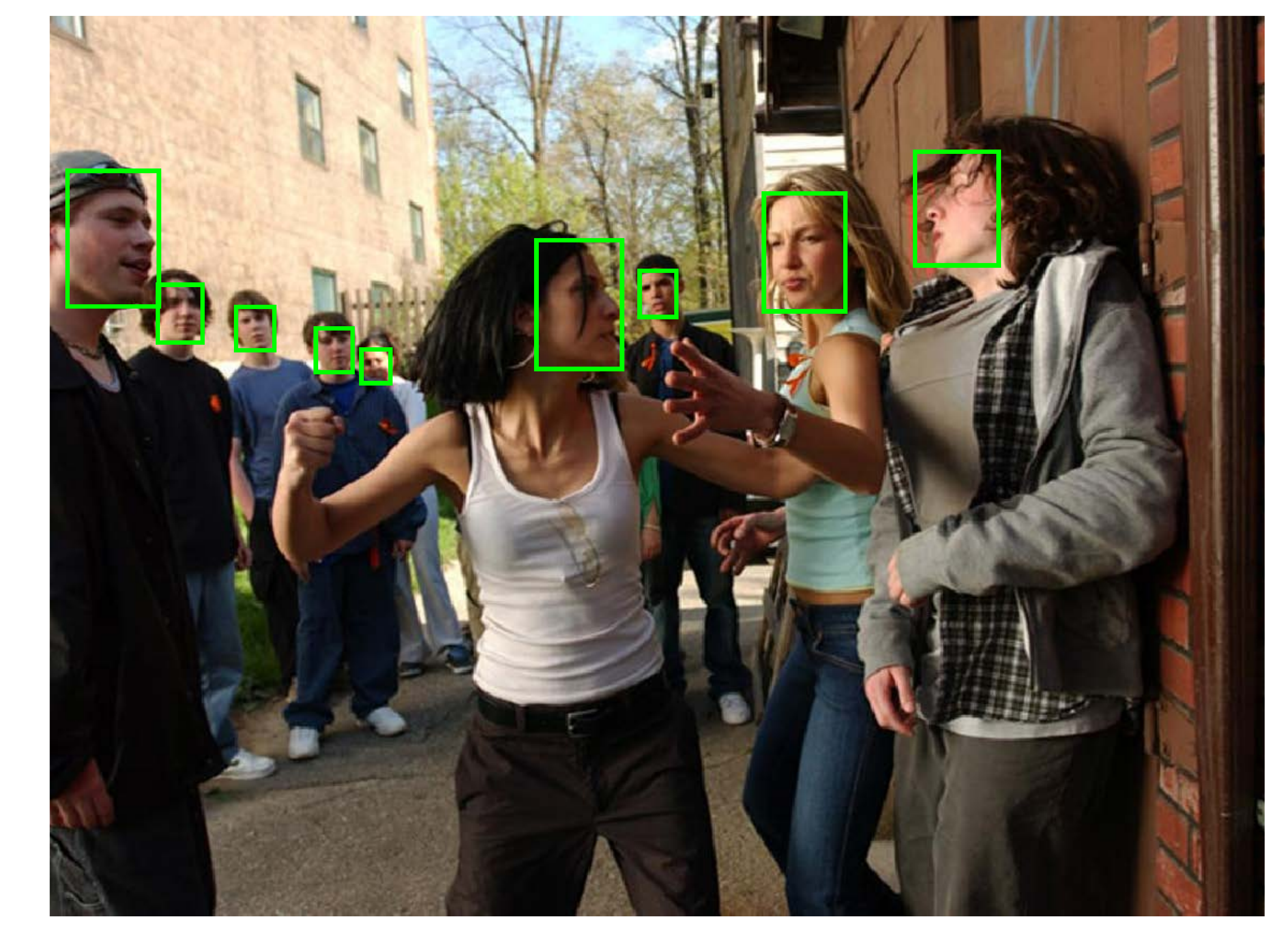}\end{minipage}&
\begin{minipage}{0.21\linewidth}\centering\includegraphics[trim={0cm 0cm 0cm 0cm},clip,width=1\linewidth,height=0.72\linewidth]{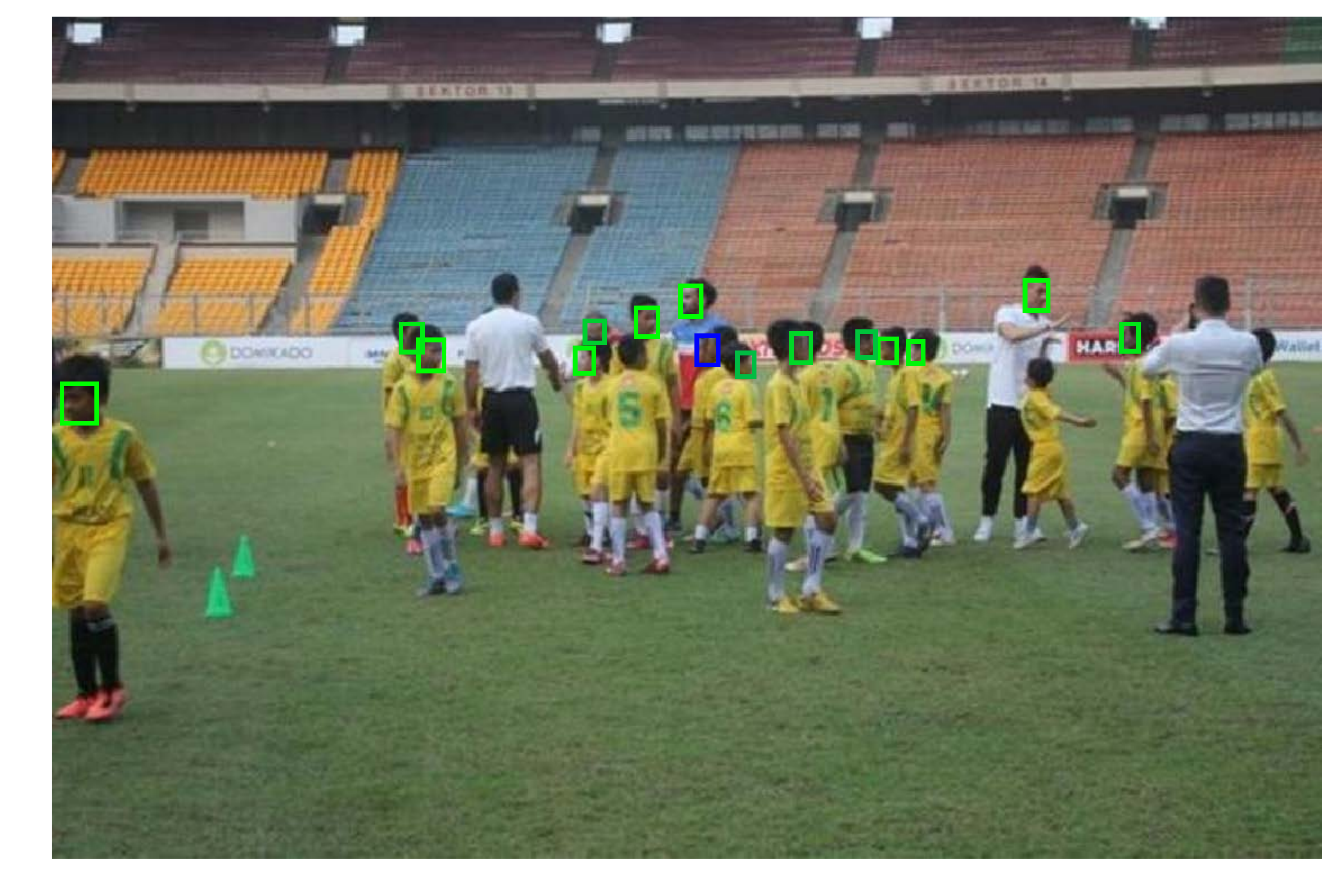}\end{minipage}&
\begin{minipage}{0.21\linewidth}\centering\includegraphics[trim={0cm 0cm 0cm 0cm},clip,width=1\linewidth,height=0.72\linewidth]{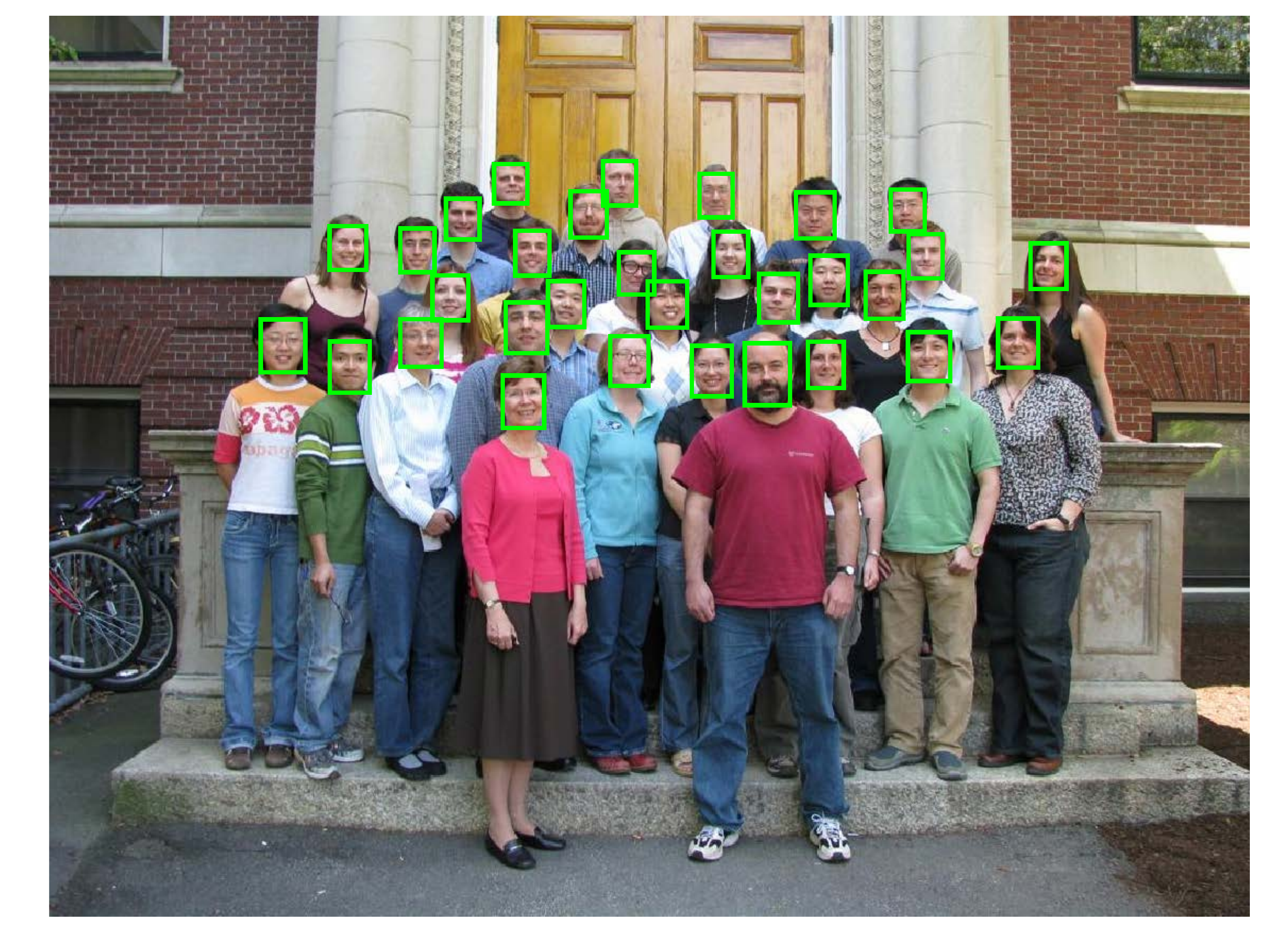}\end{minipage}&
\begin{minipage}{0.21\linewidth}\centering\includegraphics[trim={0cm 0cm 0cm 0cm},clip,width=1\linewidth,height=0.72\linewidth]{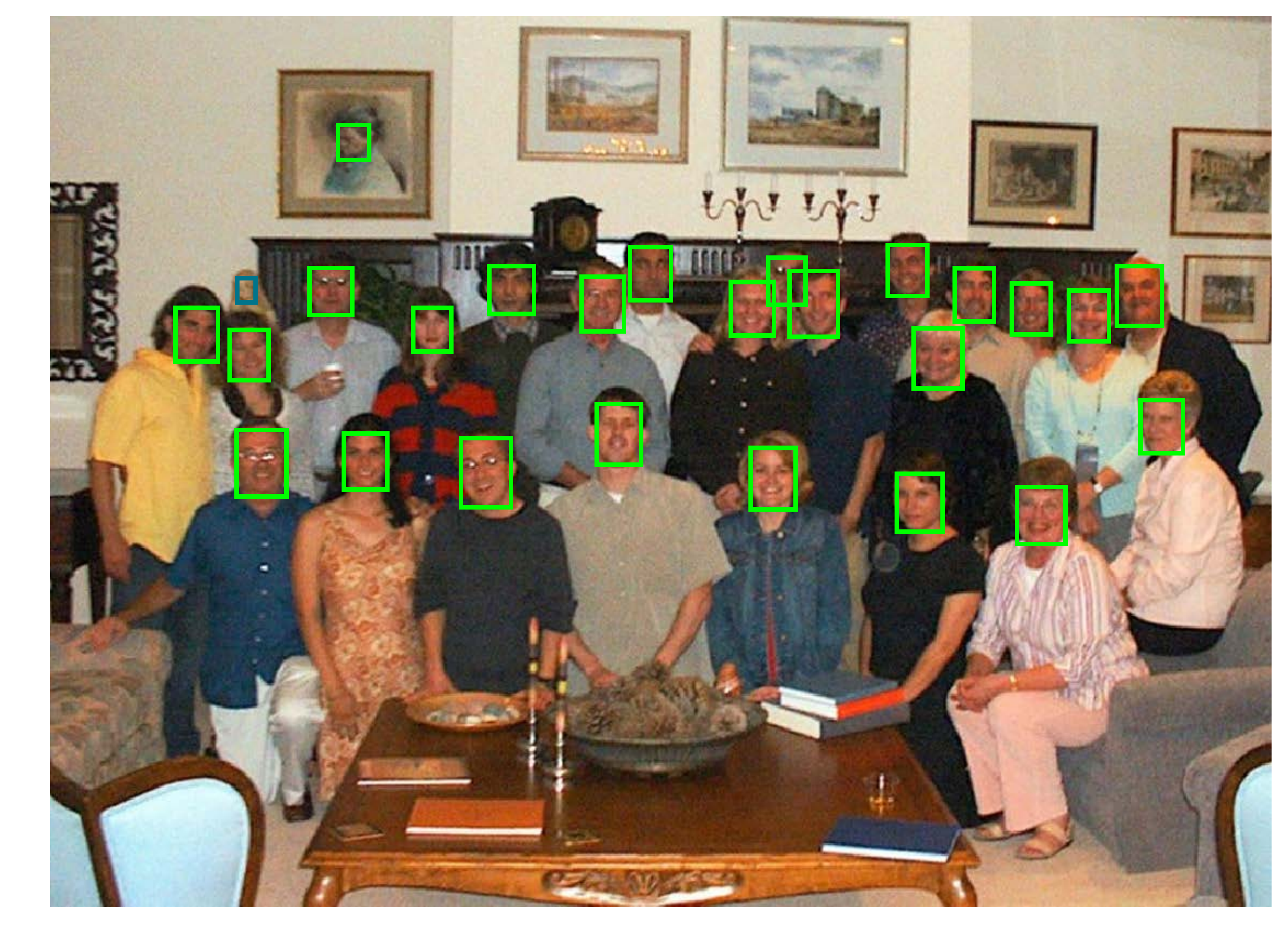}\end{minipage}\\

\begin{minipage}{0.21\linewidth}\centering\includegraphics[trim={0cm 0cm 0cm 0cm},clip,width=1\linewidth,height=0.72\linewidth]{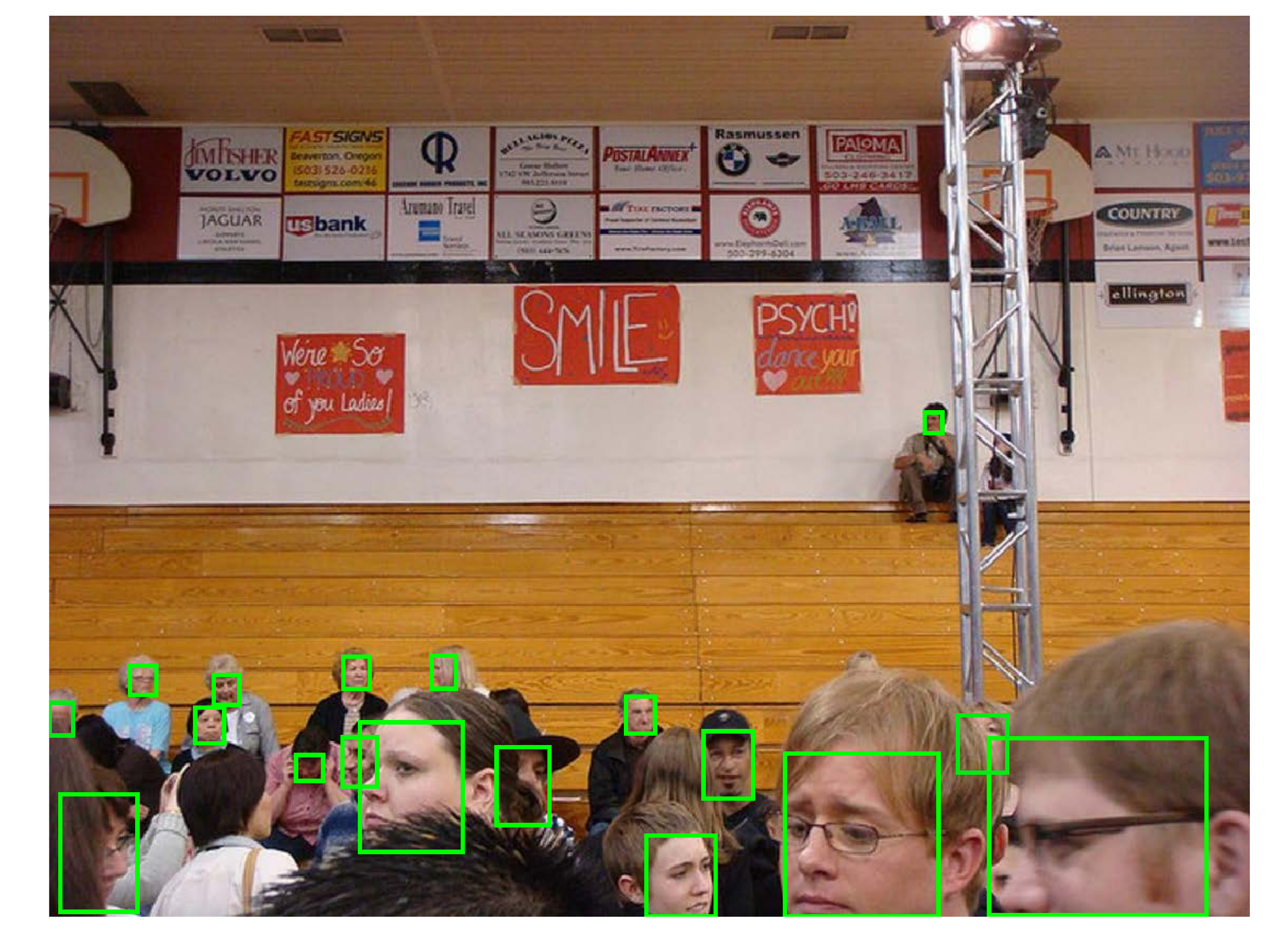}\end{minipage}&
\begin{minipage}{0.21\linewidth}\centering\includegraphics[trim={0cm 0cm 0cm 0cm},clip,width=1\linewidth,height=0.72\linewidth]{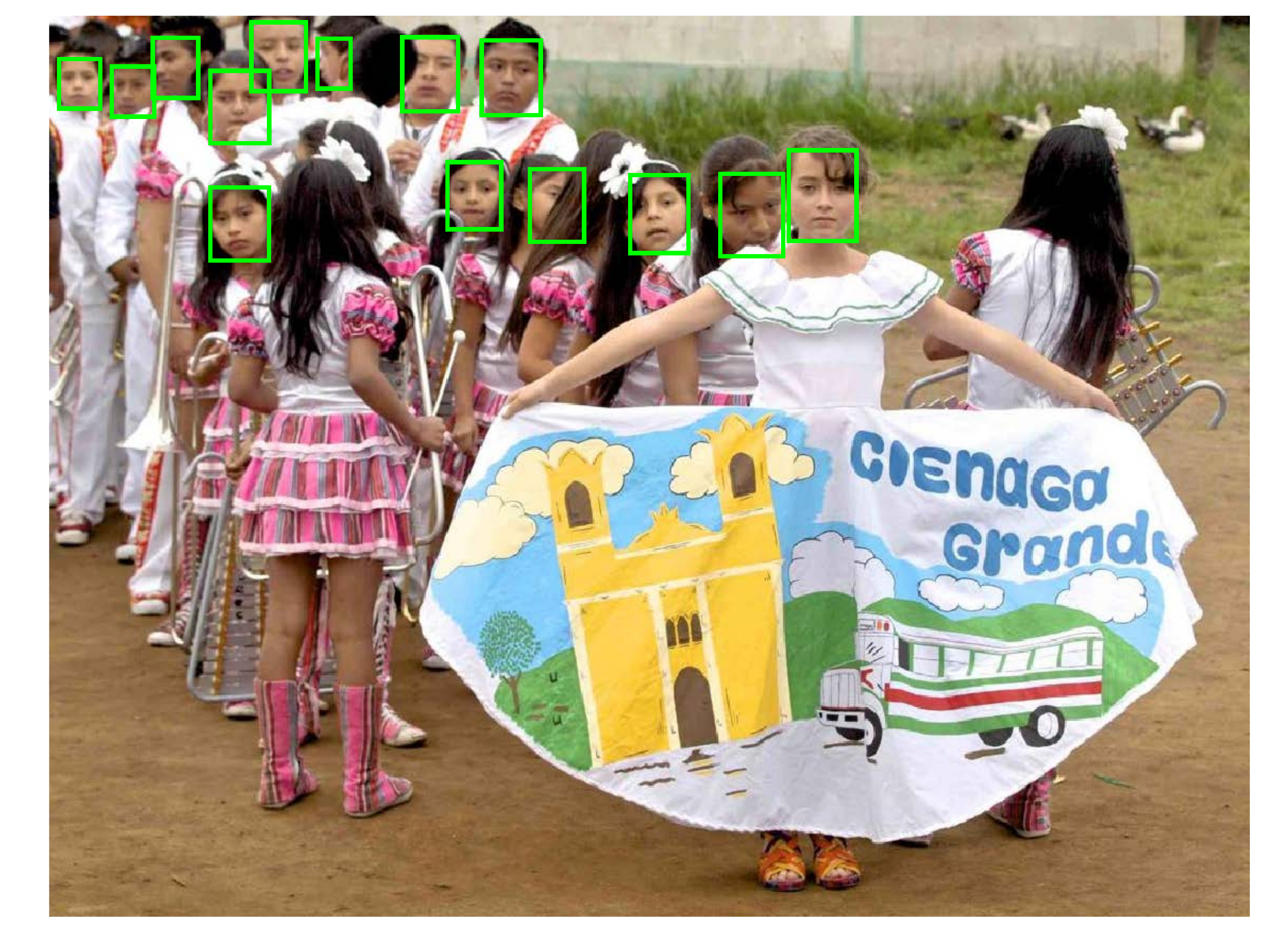}\end{minipage}&
\begin{minipage}{0.21\linewidth}\centering\includegraphics[trim={0cm 0cm 0cm 0cm},clip,width=1\linewidth,height=0.72\linewidth]{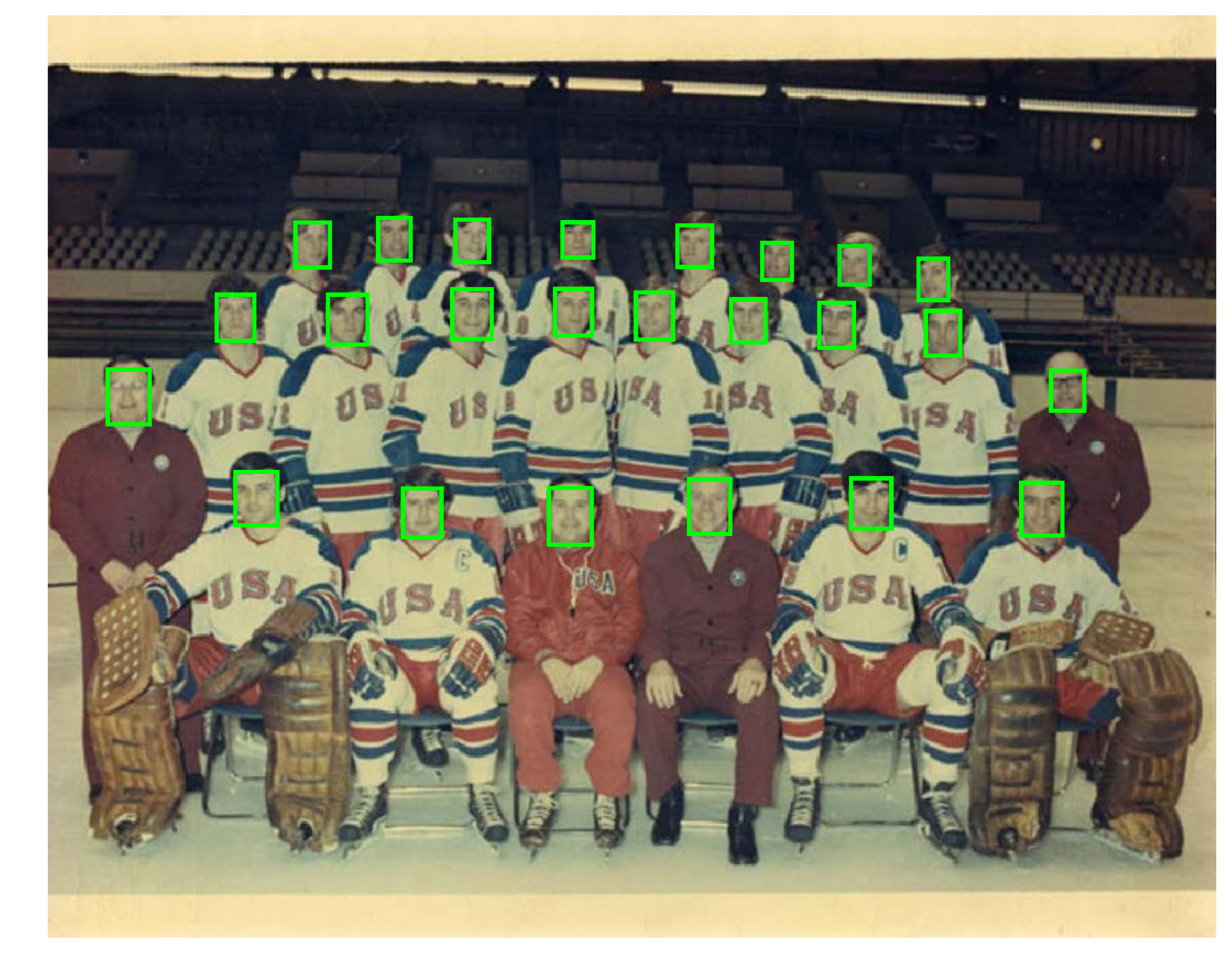}\end{minipage}&
\begin{minipage}{0.21\linewidth}\centering\includegraphics[trim={0cm 0cm 0cm 0cm},clip,width=1\linewidth,height=0.72\linewidth]{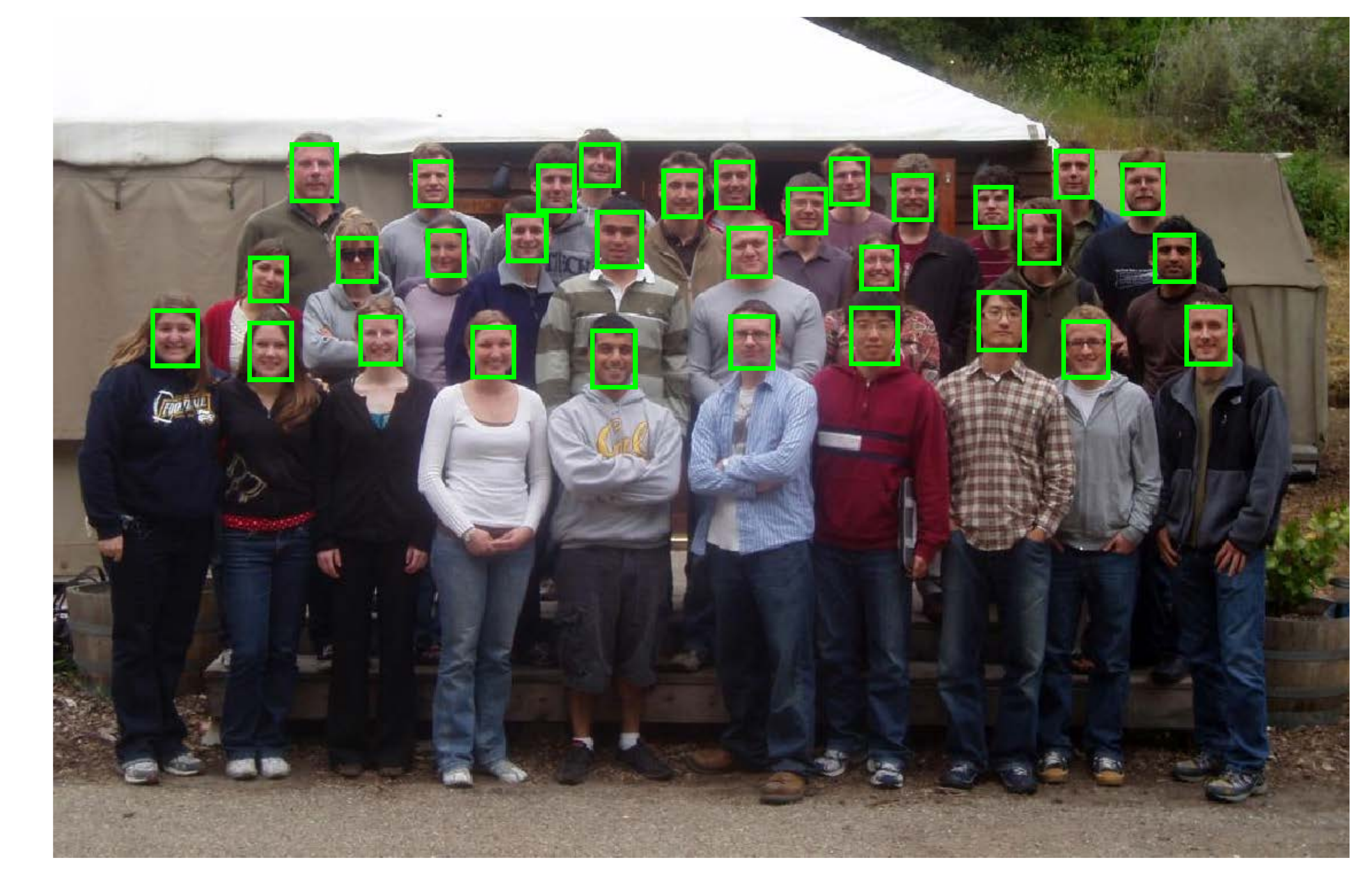}\end{minipage}\\

\hline

\end{tabular}
\caption{Qualitative results of \SSH on the validation set of the \Wider dataset.  Green and blue represent a classification score of $1.0$ and $0.5$ respectively.}
\label{tab:qual_res}
\end{figure*}

Figure \ref{tab:qual_res} shows some qualitative results on the Wider validation set. The colors encode the score of the classifier. Green and blue represent score $1.0$ and $0.5$ respectively.

\section{Conclusion}
\label{sec:conclusion}
We introduced the \SSH detector, a fast and lightweight face detector that, unlike two-stage proposal/classification approaches, detects faces in a single stage. \SSH localizes and detects faces simultaneously from the early convolutional layers in a classification network. \SSH is able to achieve state-of-the-art results without using the ``\textit{head}'' of its underlying classification network (\ie \textit{fc} layers in \VGG). Moreover, instead of processing an input pyramid, \SSH is designed to be scale-invariant while detecting different face scales in a single forward pass of the network. \SSH achieves state-of-the-art performance on the challenging \Wider dataset as well as \textit{FDDB} and \textit{Pascal-Faces} while reducing the detection time considerably.\vspace{0.3cm}

\small{\textbf{Acknowledgement} This research is based upon work supported by the Office of the Director of National Intelligence (ODNI), Intelligence Advanced Research Projects Activity (IARPA), via IARPA R\&D Contract No. 2014-14071600012. The views and conclusions contained herein are those of the authors and should not be interpreted as necessarily representing the official policies or endorsements, either expressed or implied, of the ODNI, IARPA, or the U.S. Government. The U.S. Government is authorized to reproduce and distribute reprints for Governmental purposes notwithstanding any copyright annotation thereon.}

{\small
\bibliographystyle{ieee}
\bibliography{egbib}
}

\end{document}